**Florentin Smarandache**

# Unification

# of Fusion Theories,

# Rules, Filters, Image

# Fusion and

# Target Tracking

# Methods (UFT)

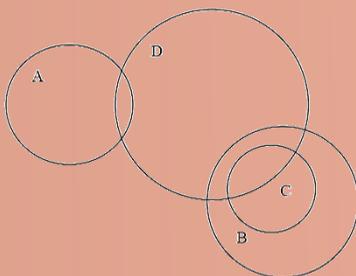

Information
Chains
Algorithms
Image
Methods
Filters
Theories

2015

Florentin Smarandache

Unification of Fusion Theories, Rules, Filters, Image
Fusion and Target Tracking Methods (UFT)

Florentin Smarandache

# Unification of Fusion Theories, Rules, Filters, Image Fusion and Target Tracking Methods (UFT)

2015

# Contents

# Foreword

Since no fusion theory neither fusion rule fully satisfy all needed applications, the author has proposed since 2004 a Unification of Fusion Theories and a Unification / Combination of Fusion Rules in solving problems/applications. For each particular application, one selects the most appropriate fusion space and fusion model, then the rules, and algorithms of implementation.

We are working in the Unification of the Fusion Theories and Rules (UFTR), which looks like a cooking recipe, better we would say like a logical chart for a computer programmer, but we do not see another method to comprise/unify all things.

The unification scenario presented herein, which is now in an incipient form, should periodically be updated incorporating new discoveries from the fusion and engineering research.

The author has pledged in various papers, conference presentations, and scientific grant applications (between 2004-2015) for the unification of fusion theories, rules, image segmentation procedures, filter algorithms, and target tracking methods for more accurate applications to our real world problems (see References below).

We want to thank for their comments to Dr. Mark Alford from Air Force Research Laboratory, RIEA, Rome, NY, USA, Dr. Wu Li from NASA Langley Research Center, the late Dr. Philippe Smets from the Université Libre de Bruxelles, Dr. Jean Dezert from ONERA in Paris, Dr. Albena Tchamova from the Bulgarian Academy of Sciences, the Russian physicist Dmitri Rabounski, and Indonesian scientist Victor Christianto.

## References related to UFT
(in chronological order)

edited by Franklin E. Hoke and Margot R. Ashcroft, <Final Technical Report>, AFRL-RI-RS-TR-2012-024, under SUNY IT Visiting Faculty Research, p. 12, January 2012.

*Presentations:*

1. F. Smarandache, An In-Depth Look at Information Fusion Rules and Unification of Fusion Theories, Invited speaker at and sponsored by NASA Langley Research Center, Hampton, VA, USA, November 5, 2004.

2. F. Smarandache, Unification of the Fusion Theories (UFT), Invited speaker at and sponsored by NATO Advance Study Institute, Albena, Bulgaria, 16-27 May, 2005.

*Grant:*

1. F. Smarandache, Unification of Fusion Theories and Rules, Filter Algorithms, and Target Tracking Methods for Applications to Air Force Problems, Faculty Grant Extension, from Information Institute, AFRL/RIB, Rome, NY, USA, 2009-2010.

# 1. Unification of Fusion Theories (UFT)

## 1.1. Introduction.

Each theory works well for some applications, but not for all. We extended the power set (from Dempster-Shafer Theory) and hyper-power set (from Dezert-Smarandache Theory) to a Boolean algebra called **super-power set** obtained by the closure of the frame of discernment under union, intersection, and complement of sets (for non-exclusive elements one considers as complement the fuzzy or neutrosophic complement). All *bba*'s (basic belief assignments) and rules are redefined on this Boolean algebra.

A similar generalization has been previously used by Guan-Bell (1993) for the Dempster-Shafer rule **using propositions in sequential logic**, herein we reconsider the Boolean algebra for all fusion rules and theories but **using sets** instead of propositions, because generally it is harder to work in sequential logic with summations and inclusions than in the set theory.

## 1.2. Fusion Space.

For n $\geq$ 2 let $\Theta = \{\theta_1, \theta_2, \ldots, \theta_n\}$ be the frame of discernment of the fusion problem/application under consideration. Then $(\Theta, \cup, \cap, \mathcal{C}), \Theta$ closed under these three operations: union, intersection, and complementation of sets respectively, forms a Boolean algebra. With respect to the partial ordering relation, the inclusion $\subseteq$, the minimum element is the empty set $\phi$, and the maximal element is the total ignorance $I = \cup_{i=1}^{n} \theta i$.

Similarly one can define: (Θ, ∪, ∩, \), for sets, Θ closed with respect to each of these operations: union, intersection, and difference of sets respectively. (Θ, ∪, ∩, $\mathcal{C}$) and (Θ, ∪, ∩, \) generate the same super-power set $S^\Theta$ closed under ∪, ∩, $\mathcal{C}$ and \ because for any $A, B \in S^\Theta$ one has $\mathcal{C}A = I \setminus A$ and reciprocally $A \setminus B = A \cap \mathcal{C}B$.

If one considers propositions, then forms a Lindenbaum algebra in sequential logic, which is isomorphic with the above (Θ, ∨, ∧, ¬) Boolean algebra.

By choosing the frame of discernment Θ closed under ∪ only, one gets DST, Yager's, TBM, DP theories. Then making Θ closed under both ∪, ∩, one gets DSm theory. While, extending Θ for closure under ∪, ∩ and $\mathcal{C}$, one also includes the complement of set (or negation of proposition if working in sequential logic); in the case of non-exclusive elements in the frame of discernment one considers a fuzzy or neutrosophic complement. Therefore the super-power set (Θ, ∪, ∩, $\mathcal{C}$) includes all the previous fusion theories.

The power set $2^\Theta$, used in DST, Yager's, TBM, DP, which is the set of all subsets of Θ, is also a Boolean algebra, closed under ∪, ∩ and $\mathcal{C}$, but does not contain intersections of elements from Θ.

The Dedekind distributive lattice $D^\Theta$ used in DSmT, is closed under ∪, ∩ and if negations/complements arise they are directly introduced in the frame of discernment, say Θ′, which is then closed under ∪, ∩.

Unlike others, DSmT allows intersections, generalizing the previous theories.

The Unifying Theory contains intersections and complements as well.

In a particular case, for a frame of discernment with two elements $\Theta = \{A, B\}$, we remark that:

a) in classical probability (Bayesian) theory:

$$P(A) + P(B) = 1, \qquad (1)$$

where $P(X)$ means probability of $X$;

b) in Dempster-Shafer Theory, which is a generalization of Bayesian theory:

$$m_{DST}(A) + m_{DST}(B) + m_{DST}(A \cup B) = 1; \qquad (2)$$

c) in Dezert-Smarandache Theory, which is a generalization of Dempster-Shafer Theory:

$$m_{DSmT}(A) + m_{DSmT}(B) + m_{DSmT}(A \cup B) +$$
$$+ m_{DSmT}(A \cap B) = 1; \qquad (3)$$

d) while in the Unification of Fusion Theory, which is a generalization of all of the above:

$$m_{UFT}(A) + m_{UFT}(B) + m_{UFT}(A \cup B) +$$
$$+ m_{UFT}(A \cap B) + m_{UFT}(\neg A) + m_{UFT}(\neg B) +$$
$$+ m_{UFT}(\neg A \cup \neg B) + m_{UFT}(\neg A \cap \neg B) = 1, \quad (4)$$

as V. Christianto observed, where $\neg X$ means the negation (or complement) of $X$.

The number of terms in the left side of this relationship is equal to: $2^{\wedge}(2^n - 1)$ and represents the number of all possible combinations of distinct parts in the Venn diagram, where $n$ is the cardinal of $\Theta$.

Let's consider a frame of discernment $\Theta$ with exclusive or non-exclusive hypotheses, exhaustive or non-exhaustive, closed or open world (all possible cases).

We need to make the remark that in case when these $n \geq 2$ elementary hypotheses $\theta_1, \theta_2, \ldots, \theta_n$ are **exhaustive and exclusive** one gets the Dempster-Shafer Theory, Yager's, Dubois-Prade Theory, Dezert-Smarandache Theory, while for the case when the hypotheses are **non-exclusive** one gets Dezert-Smarandache Theory, but for **non-exhaustivity** one gets TBM.

An exhaustive frame of discernment is called **closed world**, and a non-exhaustive frame of discernment is called **open world** (meaning that new hypotheses might exist in the frame of discernment that we are not aware of). $\Theta$ may be finite or infinite.

Let $m_j: S^\Theta \to [0, 1], 1 \le j \le s,$ (5)

be $s \ge 2$ basic belief assignments, (when *bba*'s are working with crisp numbers), or with subunitary subsets,

$$m_j: S^\Theta \to \rho[0.1],\tag{6}$$

where $\rho[0.1]$ is the set of all subsets of the interval $[0, 1]$ (when dealing with very imprecise information).

Normally, the sum of crisp masses of a *bba*, $m(.)$, is 1, i.e. $\sum_{X \in S^\wedge T} m(X) = 1.$ (7)

## 1.3. Incomplete and Paraconsistent Information.

For incomplete information the sum of a *bba* components can be less than 1 (not enough information known), while in paraconsistent information the sum can exceed 1 (overlapping contradictory information).

The masses can be normalized (i.e. getting the sum of their components = 1), or not (sum of components < 1 in incomplete information; or > 1 in paraconsistent information).

For a *bba* valued on subunitary subsets, one can consider, as normalization of $m(.)$, either:

$$\sum_{x \in S^\wedge T} sup\{m(X)\} = 1,\tag{8}$$

or that there exist crisp numbers $x \in X$ for each $X \in S^{\Theta}$ such that $\sum_{\substack{X \in S^{\wedge T} \\ x \in X}} m(x) - 1.$ (9)

Similarly, for a *bba* $m(.)$ valued on subunitary subsets dealing with paraconsistent and incomplete information respectively:

a) for incomplete information, one has

$\sum_{x \in S^{\wedge T}} sup \{m(X)\} < 1,$ (10)

b) while for paraconsistent information, one has

$\sum_{x \in S^{\wedge T}} sup \{m(X)\} > 1,$ (11)

and there do not exist crisp numbers $x \in X$ for each $X \in S^{\Theta}$ such that

$\sum_{\substack{x \in S^{\wedge T} \\ x \in X}} m(x) = 1.$ (12)

## 1.4. Specificity Chains.

We use the min principle and the precocious/prudent way of computing and transferring the conflicting mass.

Normally by transferring the conflicting mass and by normalization we diminish the specificity.

If $A \cap B$ is empty, then the mass is moved to a less specific element $A$ (also to $B$), but if we have a pessimistic view on $A$ and $B$ we move the mass $m(A \cap B)$ to $A \cup B$ (entropy increases, imprecision increases), and even more if we are very pessimistic about $A$ and $B$: we move the conflicting mass to the total ignorance in a closed world, or to the empty set in an open world.

Specificity Chains:

a) From specific to less and less specific (in a closed world):

$(A \cap B) \subset A \subset (A \cup B) \subset I,$

or $\qquad (A \cap B) \subset B \subset (A \cup B) \subset I.$ (13)

Also from specific to unknown (in an open world):

$$A \cap B \rightarrow \phi.$$ (14)

    b) And similarly for intersections of more elements:

$A \cap B \cap C$, etc.

$A \cap B \cap C \subset A \cap B \subset A \subset (A \cup B)$

$\subset (A \cup B \cup C) \subset I$ (15)

or

$(A \cap B \cap C) \subset (B \cap C) \subset B \subset (A \cup B)$

$\subset (A \cup B \cup C) \subset I$ (16)

etc. in a closed world.

    Or $A \cap B \cap C \rightarrow \phi$ in an open world. (17)

    c) Also, in a closed world:

$A \cap (B \cup C) \subset B \cup C \subset (B \cup C) \subset (A \cup B \cup C)$

$\subset I$ (18)

or

$A \cap (B \cup C) \subset A \subset (A \cup B) \subset (A \cup B \cup C)$

$\subset I$ (19)

    Or $A \cap (B \cup C) \rightarrow \phi$ in an open world. (20)

## 1.5. Static and Dynamic Fusion.

According to Wu Li from NASA we have the following classification and definitions:

- **Static fusion** means to combine all belief functions simultaneously.
- **Dynamic fusion** means that the belief functions become available one after another sequentially, and the current belief function is updated by combining itself with a newly available belief function.

## 1.6. An Algorithm (or Scenario) for the Unification of Fusion Theories.

Since everything depends on the application/problem to solve, this scenario looks like a logical chart designed by the programmer in order to write and implement a computer program, or like a cooking recipe. Here it is the scenario attempting for a unification and reconciliation of the fusion theories and rules:

1) If all sources of information are reliable, then apply the conjunctive rule, which means consensus between them (or their common part).

2) If some sources are reliable and others are not, but we don't know which ones are unreliable, apply the disjunctive rule as a cautious method (and no transfer or normalization is needed).

3) If only one source of information is reliable, but we don't know which one, then use the exclusive disjunctive rule based on the fact that $X_1 \veebar X_2 \veebar ... \veebar X_n$ means either $X_1$ is reliable, or $X_2$, or and so on, or $X_n$, but not two or more in the same time.

4) If a mixture of the previous three cases, in any possible way, use the mixed conjunctive-disjunctive rule.

Suppose we have four sources of information and we know that: either the first two are telling the truth, or the third, or the fourth. The mixed formula becomes:

$$m_{\cap \cup}(\phi) = 0, \text{ and } \forall A \in S^{\Theta} \setminus \phi, \text{ one has} \qquad (21)$$

$$m_{\cap \cup}(A) = \sum_{\substack{X_1, X_2, X_3, X_4 \in S^{\wedge \Theta} \\ ((X_1 \cap X_2) \cup X_3) e \cup X_4 = A}} m_1(X_1) m_2(X_2) m_3(X_3) m_4(X_4).$$

5) If we know the sources which are unreliable, we discount them. But if all sources are fully unreliable (100%), then the fusion result becomes vacuum *bba* (i.e. $m(\Theta) = 1$, and the problem is indeterminate. We need to get new sources which are reliable or at least they are not fully unreliable.

6) If all sources are reliable, or the unreliable sources have been discounted (in the **default case**), then use the DSm classic rule (which is commutative, associative, Markovian) on Boolean algebra $(\Theta, \cup, \cap, \mathcal{C})$, no matter what contradictions (or model) the problem has. I emphasize that the super-power set $S^\Theta$ generated by this Boolean algebra contains singletons, unions, intersections, and complements of sets.

7) If the sources are considered from a statistical point of view, use **Murphy's average rule** (and no transfer or normalization is needed).

8) In the case the model is not known (the default case), it is prudent/cautious to use the free model (i.e. all intersections between the elements of the frame of discernment are nonempty) and DSm classic rule on $S^\Theta$, and later if the model is found out (i.e. the constraints of empty intersections become known), one can adjust the conflicting mass at any time/moment using the DSm hybrid rule.

9) Now suppose the model becomes known [i.e. we find out about the contradictions (=empty intersections) or consensus (=non-empty intersections) of the problem/application]. Then:

9.1) If an intersection $(A \cap B)$ is not empty, we keep the mass $m(A \cap B)$ on $(A \cap B)$, which means consensus

(common part) between the two hypotheses $A$ and $B$ (i.e. both hypotheses $A$ and $B$ are right) [here one gets DSmT].

9.2) If the intersection $(A \cap B) = \phi$ is empty, meaning contradiction, we do the following:

9.2.1) if one knows that between these two hypotheses $A$ and $B$ one is right and the other is false, but we don't know which one, then one transfers the mass $m(A \cap B)$ to $(A \cup B)$, since $A \cup B$ means at least one is right [here one gets *Yager's* if $n = 2$, or *Dubois-Prade*, or *DSmT*];

9.2.2) if one knows that between these two hypotheses $A$ and $B$ one is right and the other is false, and we know which one is right, say hypothesis $A$ is right and $B$ is false, then one transfers the whole mass $m(A \cap B)$ to hypothesis $A$ (nothing is transferred to $B$);

9.2.3) if we don't know much about them, but one has an optimistic view on hypotheses $A$ and $B$, then one transfers the conflicting mass $m(A \cap B)$ to $A$ and $B$ (the nearest specific sets in the Specificity Chains) [using *Dempster's, PCR2-5*];

9.2.4) if we don't know much about them, but one has a pessimistic view on hypotheses $A$ and $B$, then one transfers the conflicting mass $m(A \cap B)$ to $(A \cup B)$ (the more pessimistic the further one gets in the Specificity Chains: $(A \cap B) \subset A \subset (A \cup B) \subset I$; this is also the **default case** [using *DP's, DSm hybrid rule, Yager's*]; if one has a very pessimistic view on hypotheses $A$ and $B$, then one transfers the conflicting mass $m(A \cap B)$ to the total ignorance in a closed world [*Yager's, DSmT*], or to the empty set in an open world [*TBM*];

9.2.5.1) if one considers that no hypothesis between $A$ and $B$ is right, then one transfers the mass $m(A \cap B)$ to other non-empty sets (in the case more hypotheses do exist in the frame of discernment) - different from $A$, $B$, $A \cup B$ - for the reason that: if $A$ and $B$ are not right then there is a bigger chance that other hypotheses in the frame of discernment have a higher subjective probability to occur; we do this transfer in a **closed world** [DSm hybrid rule]; but, if it is an **open world**, we can transfer the mass $m(A \cap B)$ to the empty set leaving room for new possible hypotheses [here one gets TBM];

9.2.5.2) if one considers that none of the hypotheses $A$, $B$ is right and no other hypothesis exists in the frame of discernment (i.e. $n = 2$ is the size of the frame of discernment), then one considers the **open world** and one transfers the mass to the empty set [here DSmT and TBM converge to each other].

Of course, this procedure is extended for any intersections of two or more sets: $A \cap B \cap C$, etc. and even for mixed sets: $A \cap (B \cup C)$, etc.

If it is a dynamic fusion in a real time and associativity and/or Markovian process are needed, use an algorithm which transforms a rule (which is based on the conjunctive rule the previous result of the conjunctive rule and, depending of the rule, other data. Such rules are called quasi-associative and quasi-Markovian.

Some applications require the necessity of **decaying the old sources** because their information is considered to be worn out.

If some *bba* is not normalized (i.e. the sum of its components is $< 1$ as in incomplete information, or $> 1$ as in paraconsistent information) we can easily divide each component by the sum of the components and normalize it. But also it is possible to fusion incomplete and paraconsistent masses, and then normalize them after fusion. Or leave them unnormalized since they are incomplete or paraconsistent.

PCR5 does the most mathematically exact (in the fusion literature) redistribution of the conflicting mass to the elements involved in the conflict, redistribution that exactly follows the tracks of the conjunctive rule.

## 1.7. Examples.

## 1.7.1. Bayesian Example.

Let $\Theta = \{A, B, C, D, E\}$ be the frame of discernment.

| | A | B | C | D | E | A∩B | A∩C | A∩D | A∩E | B∩C | B∩D |
|---|---|---|---|---|---|---|---|---|---|---|---|
| | | | | | | ≠ | = | = | = | Not known if =or ≠ | = |
| | | | | | | φ | φ | φ | φ | φ | φ |
| | | | | | | Consensus between A and B | Contradiction between A and C, but optimistic in both of them | One right, one wrong, but don't know which one | A is right, B is wrong | Don't know the exact model | Unknown any relation between B and D |
| $m_1$ | 0.2 | 0 | 0.3 | 0.4 | 0.1 | | | | | | |
| $m_2$ | 0.5 | 0.2 | 0.1 | 0 | 0.2 | | | | | | |
| $m_{12}$ | 0.10 | 0 | 0.03 | 0 | 0.02 | 0.04 | 0.17 | 0.20 | 0.09 | 0.06 | 0.08 |
| | | | | | | ↓ | ↓ | ↓ | ↓ | ↓ | ↓ |
| | | | | | | A∩B | A, C | A∪B | A | B∩C We keep the mass 0.06 on B∩C till we find out more information on the model. | B∪D |
| $m_e$ | | | | | | 0.04 | 0.107, 0.063 | 0.20 | 0.09 | 0.06 | 0.08 |
| $m_{UFT}$ | 0.324 | 0.040 | 0.119 | 0 | 0.027 | 0.04 | 0 | 0 | 0 | 0.06 | 0 |
| $m_{lower}$ | 0.10 | 0 | 0.03 | 0 | 0.02 | | | | | | |

| | | | | | |
|---|---|---|---|---|---|
| (closed world) $m_{lower}$ (open world) | 0.10 | 0 | 0.03 | 0 | 0.02 |
| $m_{middle}$ (default) | 0.10 | 0 | 0.03 | 0 | 0.02 |
| $m_{upper}$ | 0.400 | 0.084 | 0.178 | 0.227 | 0.111 |

*Table 1. Part 1.* Bayesian Example using the Unified Fusion Theories rule regarding a mixed redistribution of partial conflicting masses.

| | B∩E | C∩D | C∩E | D∩E | A∪B | A∪C | A∪D | A∪E | B∪C |
|---|---|---|---|---|---|---|---|---|---|
| | ≢ φ | = φ | = φ | = φ | | | | | |
| | The intersection is not empty, but neither B∩E nor B∪E interest us | Pessimistic in both C and D | Very pessimistic in both C and E | Both D and E are wrong | | | | | |
| $m_1$ | | | | | | | | | |
| $m_2$ | | | | | | | | | |
| $m_{12}$ | 0.02 | 0.04 | 0.07 | 0.08 | | | | | |
| | ↓ B, E | ↓ C∪D | ↓ A∪B∪C∪D∪E | ↓ A,B,C | | | | | |
| $m_r$ | 0.013, 0.007 | 0.04 | 0.07 | 0.027, 0.027, 0.027 | | | | | |
| $m_{UFT}$ | 0 | 0 | 0 | 0 | 0 | 0 | 0.20 | 0 | 0 |
| $m_{lower}$ (closed world) | | | | | | | | | |
| $m_{lower}$ (open world) | | | | | | | | | |
| $m_{middle}$ (default) | | | | | 0.04 | 0.17 | 0.20 | 0.09 | 0.06 |
| $m_{upper}$ | | | | | | | | | |

*Table 1. Part 2.* Bayesian Example using the Unified Fusion Theories rule regarding a mixed redistribution of partial conflicting masses.

| | B∪D | B∪E | C∪D | C∪E | D∪E | A∪B∪C∪D∪E | φ |
|---|---|---|---|---|---|---|---|
| | | | | | | | |
| | | | | | | | |
| $m_1$ | | | | | | | |
| $m_2$ | | | | | | | |
| $m_{12}$ | | | | | | | |
| | | | | | | | |
| $m_r$ | | | | | | | |
| $m_{UFT}$ | 0.08 | 0 | 0.04 | 0 | 0 | 0.07 | 0 |
| $m_{lower}$ (closed world) | | | | | | 0.85 | |
| $m_{lower}$ (open world) | | | | | | | 0.85 |
| $m_{middle}$ (default) | 0.08 | 0.02 | 0.04 | 0.07 | 0.08 | | |
| $m_{upper}$ | | | | | | | |

*Table 1. Part 3.* Bayesian Example using the Unified Fusion Theories rule regarding a mixed redistribution of partial conflicting masses.

We keep the mass $m_{12}(B \cap C) = 0.06$ on $B \cap C$ (eleventh column in *Table 1. Part 1*) although we do not know if the intersection $B \cap C$ is empty or not (this is considered the default model), since in the case when it is empty one considers an open world because $m_{12}(\phi) = 0.06$, meaning that there might be new possible hypotheses in the frame of discernment, but if $B \cap C \neq \phi$ one considers a consensus between $B$ and $C$. Later, when finding out more information about the relation between $B$ and $C$, one can transfer the mass 0.06 to $B \cup C$, or to the total ignorance $I$, or split it between the elements $B$, $C$, or even keep it on $B \cap C$.

$m_{12}(A \cap C) = 0.17$ is redistributed to A and C using the PCR5:

$$a1/0.2 = c1/0.1 = 0.02(0.2 + 0.1),$$
whence $$a1 = 0.2(0.02/0.3) = 0.013,$$
$$c1 = 0.1(0.02/0.3) = 0.007.$$
$$a2/0.5 = c2/0.3 = 0.15(0.5 + 0.3),$$
whence $$a2 = 0.5(0.15/0.8) = 0.094,$$
$$c2 = 0.3(0.15/0.8) = 0.056.$$

Thus $A$ gains:
$$a1 + a2 = 0.013 + 0.0094 = 0.107$$
and $C$ gains:

$$c1 + c2 = 0.007 + 0.056 = 0.063$$

$m_{12}(B \cap E) = 0.02$ is redistributed to $B$ and $E$ using the PCR5:

$$b/0.2 = e/0.1 = 0.02/(0.2 + 0.1),$$

whence
$$b = 0.2(0.02/0.3) = 0.013,$$
$$e = 0.1(0.02/0.3) = 0.007.$$

Thus $B$ gains 0.013 and $E$ gains 0.007.

Then one sums the masses of the conjunctive rule $m_{12}$ and the redistribution of conflicting masses $m_r$ (according to the information we have on each intersection, model, and relationship between conflicting hypotheses) in order to get the mass of the Unification of Fusion Theories rule $m_{UFT}$.

$m_{UFT}$, the Unification of Fusion Theories rule, is a combination of many rules and gives the optimal redistribution of the conflicting mass for each particular problem, following the given model and relationships between hypotheses; this extra-information allows the choice of the combination rule to be used for each intersection. The algorithm is presented above.

$m_{lower}$, the lower bound believe assignment, the most pessimistic/prudent belief, is obtained by transferring the whole conflicting mass to the total ignorance (Yager's rule) in a closed world, or to the empty set (Smets' TBM) in an open world herein meaning that other hypotheses might belong to the frame of discernment.

$m_{middle}$, the middle believe assignment, half optimistic and half pessimistic, is obtained by transferring the partial conflicting masses $m_{12}(X \cap Y)$ to the partial ignorance $X \cup Y$ (as in Dubois-Prade theory or more general as in Dezert-Smarandache theory). Another way to compute a middle believe assignment would be to average the $m_{lower}$ and $m_{upper}$.

$m_{upper}$, the upper bound believe assignment, the most optimistic (less prudent) belief, is obtained by transferring the masses of intersections (empty or non-empty) to the elements in the frame of discernment using the PCR5 rule of combination, i.e. $m_{12}(X \cap Y)$ is split to the elements $X, Y$ (see *Table 2*). We use PCR5 because it is more exact mathematically (following backwards the tracks of the conjunctive rule) than Dempster's rule, minC, and PCR1-4.

| X | $m_{12}(X)$ | A | B | C | D | E |
|---|---|---|---|---|---|---|
| A∩B | 0.040 | 0.020 | 0.020 | | | |
| A∩C | 0.170 | 0.107 | | 0.063 | | |
| A∩D | 0.200 | 0.111 | | | 0.089 | |
| A∩E | 0.090 | 0.020 | | | | 0.020 |
| | | 0.042 | | | | 0.008 |
| B∩C | 0.060 | | 0.024 | 0.036 | | |
| B∩D | 0.080 | | 0.027 | | 0.053 | |
| B∩E | 0.020 | | 0.013 | | | 0.007 |
| C∩D | 0.040 | | | 0.008 | 0.032 | |
| C∩E | 0.070 | | | 0.036 | | 0.024 |
| | | | | 0.005 | | 0.005 |
| D∩E | 0.080 | | | | 0.053 | 0.027 |
| Total | 0.850 | 0.300 | 0.084 | 0.148 | 0.227 | 0.091 |

*Table 2.* Redistribution of the intersection masses to the singletons A, B, C, D, E using the PCR5 rule only, needed to compute the upper bound belief assignment $m_{upper}$.

## 1.7.2. Negation/Complement Example.

Let $\Theta = \{A, B, C, D\}$ be the frame of discernment. Since $(\Theta, \cup, \cap, \mathcal{C})$ is Boolean algebra, the super-power set $S^\theta$ includes complements/negations, intersections and unions. Let's note by $\mathcal{C}(B)$ the complement of B.

| | A | B | D | $\mathcal{C}(B)$ | A∩C {Later in the dynamic fusion process we find out that A∩C is empty} | B∪C= B | A∩B | A∩$\mathcal{C}$(B)=A |
|---|---|---|---|---|---|---|---|---|
| | | | | | = | | = | ≠ |
| | | | | | φ | | φ | φ |
| | | | | | Unknown relationship between A and C | | Optimistic in both A and B. | Consensus between A and $\mathcal{C}$(B), but A⊆$\mathcal{C}$(B) |
| $m_1$ | 0.2 | 0.3 | 0 | 0.1 | 0.1 | 0.3 | | |
| $m_2$ | 0.4 | 0.1 | 0 | 0.2 | 0.2 | 0.1 | | |
| $m_{12}$ | 0.08 | 0.09 | 0 | 0.02 | 0.17 | 0.03 | 0.14 | 0.08 |
| | | | | | ↓ A∪C | ↓ B | ↓ A,B | ↓ A |
| $m_r$ | | | | | 0.17 | 0.03 | 0.082, 0.058 | 0.08 |
| $m_{UFT}$ | 0.277 | 0.318 | 0.035 | 0.020 | 0 | 0 | 0 | 0 |
| $m_{lower}$ (closed world) | 0.16 | 0.26 | 0 | 0.02 | 0 | 0 | | |
| $m_{lower}$ (open world) | 0.16 | 0.26 | 0 | 0.02 | 0 | 0 | | |
| $m_{middle}$ (default) | 0.16 | 0.26 | 0 | 0.02 | 0 | 0 | | |
| $m_{upper}$ | 0.296 | 0.230 | 0 | 0.126 | 0.219 | 0.129 | | |

*Table 3. Part 1.* Negation/Complement Example using the Unified Fusion Theories rule regarding a mixed redistribution of partial conflicting masses.

| | A∩ (B∪C) | B∩ $\mathcal{C}$(B) | B∩(A∩C) | $\mathcal{C}$(B)∩(A∩C) | $\mathcal{C}$(B)∩(B∪C) = $\mathcal{C}$(B)∩C | B∪(A∩C) =B |
|---|---|---|---|---|---|---|
| | = | = | = | = | = | |
| | φ | φ | φ | φ | φ | |
| | At least one is right between A and B∪C | B is right, $\mathcal{C}$(B) is wrong | No relationship known between B and A∩C (default case) | Very pessimistic on $\mathcal{C}$(B) and A∩C | Neither $\mathcal{C}$(B) nor B∪C are right | |
| $m_1$ | | | | | | |
| $m_2$ | | | | | | |
| $m_{12}$ | 0.14 | 0.07 | 0.07 | 0.04 | 0.07 | |
| | ↓ A∪(B∪C) | ↓ B | ↓ B∪(A∩C)=B | ↓ A∪B∪C∪D | ↓ A, D | ↓ B |
| $m_r$ | 0.14 | 0.07 | 0.07 | 0.04 | 0.035, 0.035 | |
| $m_{UFT}$ | 0 | 0 | 0 | 0 | 0 | 0 |
| $m_{lower}$ (closed world) | | | | | | |
| $m_{lower}$ (open world) | | | | | | |
| $m_{middle}$ (default) | | | | | | |
| $m_{upper}$ | | | | | | |

*Table 3. Part 2.* Negation/Complement Example using the Unified Fusion Theories rule regarding a mixed redistribution of partial conflicting masses.

| | A∪B | A∪C | A∪D | B∪C | B∪D | C∪D | A∪B∪C | A∪B∪C∪D | φ |
|---|---|---|---|---|---|---|---|---|---|
| | | | | | | | | | |
| | | | | | | | | | |
| $m_1$ | | | | | | | | | |
| $m_2$ | | | | | | | | | |
| $m_{12}$ | | | | | | | | | |
| | | | | | | | | | |
| $m_r$ | | | | | | | | | |
| $m_{UFT}$ | 0 | 0.170 | 0 | 0 | 0 | 0 | 0.140 | 0.040 | 0 |
| $m_{lower}$ (closed world) | | | | | | | | 0.56 | |
| $m_{lower}$ (open world) | | | | | | | | | 0.56 |
| $m_{middle}$ (default) | 0.14 | 0.17 | | 0.03 | | | 0.14 | 0.11 | |
| $m_{upper}$ | | | | | | | | | |

*Table 3. Part 3.* Negation/Complement Example using the Unified Fusion Theories rule regarding a mixed redistribution of partial conflicting masses.

*Model of Negation/Complement Example.*

$$A \cap B = \phi, C \subset B, A \subset \mathcal{C}(B).$$

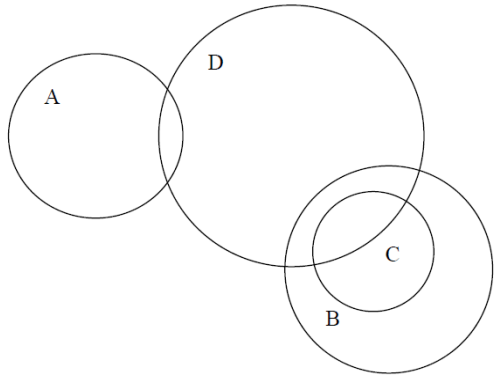

*Figure 1.*

$$m_{12}(A \cap B) = 0.14.$$
$$x1/0.2 = y1/0.1 = 0.02/0.3,$$
whence $\quad x1 = 0.2(0.02/0.3) = 0.013,$
$$y1 = 0.1(0.02/0.3) = 0.007;$$

$$x2/0.4 \ = \ y2/0.3 \ = \ 0.12/0.7,$$

whence $\qquad x2 \ = \ 0.4(0.12/0.7) \ = \ 0.069,$

$$y2 \ = \ 0.3(0.12/0.7) \ = \ 0.051.$$

Thus, $A$ gains:

0.013+0.069=0.082

and $B$ gains

$0.007 + 0.051 = 0.058.$

For the upper belief assignment $m_{upper}$ one considered all resulted intersections from results of the conjunctive rule as empty and one transferred the partial conflicting masses to the elements involved in the conflict using PCR5.

All elements in the frame of discernment were considered non-empty.

### 1.7.3. Example with Intersection.

Look at this:

Suppose

$A = \{x < 0.4\}, \ B = \{0.3 < x < 0.6\}, \ C = \{x > 0.8\} \ $.

The frame of discernment $T = \{A, B, C\}$ represents the possible cross section of a target, and there are two sensors giving the following *bbas*:

$$m_1(A) = 0.5, \ m_1(B) = 0.2, \ m_1(C) = 0.3;$$
$$m_2(A) = 0.4, \ m_2(B) = 0.4, \ m_2(C) = 0.2.$$

| | A | B | C | A∩B= {.3<x<.4} | A∪C | B∪C |
|---|---|---|---|---|---|---|
| $m_1$ | .5 | .2 | .3 | | | |
| $m_2$ | .4 | .4 | .2 | | | |
| $m_1$&$m_2$ DSmT | .20 | .08 | .06 | .28 | .22 | .16 |

*Table 4.*

We have a DSm hybrid model (one intersection A&B=nonempty). This example proves the necessity of allowing intersections of elements in the frame of discernment. [Shafer's model does not apply here.] Dezert-Smarandache Theory of Uncertain and Paradoxist Reasoning (DSmT) is the only theory that accepts intersections of elements.

### 1.7.4. Another Multi-Example of UFT.

Cases:

1. Both sources reliable: use conjunctive rule [default case]:

  1.1. $A \cap B = \phi$:

   1.1.1. Consensus between A and B; mass $\rightarrow A \cap B$;

   1.1.2. Neither $A \cap B$ nor $A \cup B$ interest us; mass $\rightarrow A, B$;

  1.2. $A \cap B = \phi$:

   1.2.1. Contradiction between $A$ and $B$, but optimistic in both of them; mass $\rightarrow A, B$;

   1.2.2. One right, one wrong, but don't know which one; mass $\rightarrow A \cup B$;

   1.2.3. Unknown any relation between $A$ and $B$ [default case]; mass $\rightarrow A \cup B$;

   1.2.4. Pessimistic in both $A$ and $B$; mass $\rightarrow A \cup B$;

   1.2.5. Very pessimistic in both $A$ and $B$;

   1.2.5.1. Total ignorance $\supset A \cup B$ mass $\rightarrow A \cup B \cup C \cup D$ (total ignorance);

1.2.5.2. Total ignorance $\rightarrow A \cup B$; mass $\phi$ (open world);

1.2.6. $A$ is right, $B$ is wrong; mass $\rightarrow A$;

1.2.7. Both $A$ and $B$ are wrong; mass $\rightarrow C, D$;

1.3. Don't know if $A \cap B =$ or $\neq \phi$ (don't know the exact model); mass $\rightarrow A \cap B$ (keep the mass on intersection till we find out more info) [default case];

2. One source reliable, other not, but not known which one: use disjunctive rule; no normalization needed.

3. $S_1$ reliable, $S_2$ not reliable 20%: discount $S_2$ for 20% and use conjunctive rule.

| | A | B | A∪B | A∩B | φ (open world) | A∪B∪C∪D | C | D |
|---|---|---|---|---|---|---|---|---|
| S1 | .2 | .5 | .3 | | | | | |
| S2 | .4 | .4 | .2 | | | | | |
| S1&S2 | .24 | .42 | .06 | .28 | | | | |
| S1 or S2 | .08 | .20 | .72 | 0 | | | | |
| UFT 1.1.1 | .24 | .42 | .06 | .28 | | | | |
| UFT 1.1.2 (PCR5) | .356 | .584 | .060 | 0 | | | | |
| UFT 1.2.1 | .356 | .584 | .060 | 0 | | | | |
| UFT 1.2.2 | .24 | .42 | .34 | 0 | | | | |
| UFT 1.2.3 | .24 | .42 | .34 | 0 | | | | |
| UFT 1.2.4 | .24 | .42 | .34 | 0 | | | | |
| UFT 1.2.5.1 | .24 | .42 | .06 | 0 | 0 | .28 | | |
| UFT 1.2.5.2 | .24 | .42 | .06 | 0 | .28 | | | |
| 80% S2 | .32 | .32 | .16 | | | .20 | | |
| UFT 1.2.6 | .52 | .42 | .06 | | | | | |
| UFT 1.2.7 | .24 | .42 | .06 | 0 | | | .14 | .14 |
| UFT 1.3 | .24 | .42 | .06 | .28 | | | | |
| UFT 2 | .08 | .20 | .72 | 0 | | | | |
| UFT 3 | .232 | .436 | .108 | .224 | | 0 | | |

*Table 5.*

# 2. Unification of Fusion Rules (UFR)

We now give a formula for the unification of a class of fusion rules based on the conjunctive and/or disjunctive rule at the first step, and afterwards the redistribution of the conflicting and/or non-conflicting mass to the non-empty sets at the second step.

Fusion of masses $m_1(.)$ and $m_2(.)$ is done directly proportional with some parameters and inversely proportional with other parameters (parameters that the hypotheses depend upon). We denote the resulting mass by $m_{UFR}(.)$.

   a) If variable $y$ is **directly proportional** with variable $p$, then $y = k_1 \cdot p$, where $k_1 \neq 0$ is a constant.
   b) If variable $y$ is inversely proportional with variable $q$, then $y = k_2 \cdot (1/q)$, where $k_2 \neq 0$ is a constant; we can also say herein that $y$ is directly proportional with variable $1/q$.

In a general way, we say that if $y$ is directly proportional with variables $p_1, p_2, ..., p_m$ and inversely proportionally with variables $q_1, q_2, ..., q_n$, then:

$$y = k \cdot \frac{(p_1 \cdot p_2 \cdot ... \cdot p_m)}{(q_1 \cdot q_2 \cdot ... \cdot q_n)} = k \cdot P/Q, \qquad (22)$$

where

$$P = \prod_{i=1}^{m} p_i, \quad Q = \prod_{j=1}^{n} q_j,$$

and $k \neq 0$ is a constant.

With such notations, we have a general formula for a UFR rule:

$m_{\mathrm{UFR}}(\varphi) = 0$, and $\forall\, A \in S^{\theta}\backslash\varphi$ one has:

$$m_{UFR}(A) = \sum_{\substack{X_1, X_2 \in S^{\wedge}\Theta \\ X_1 * X_2 = A}} d(X_1 * X_2) T(X_1, X_2)$$

$$+ \frac{P(A)}{Q(A)} \sum_{\substack{X \in S^{\wedge}\Theta \backslash A \\ X^* A \in Tr}} d(X * A) \frac{T(A, X)}{P(A)/Q(A) + P(X)/Q(X)} \tag{23}$$

where:

- $*$ can be an intersection or a union of sets;
- $d(X * Y)$ is the degree of intersection or union;
- $T(X, Y)$ is a T-norm/conorm in fuzzy set/logic (or N-norm/conorm in a more general case, in neutrosophic set/logic) class of fusion combination rules respectively (extension of conjunctive or disjunctive rules respectively to fuzzy or neutrosophic operators) or any other fusion rule; the T-norm and N-norm correspond to the intersection of sets, while the T-conorm and N-conorm to the disjunction of sets;
- $Tr$ is the ensemble of sets (in majority cases they are empty sets) whose masses must be transferred (in majority cases to non-empty sets, but there are exceptions for the open world);"
- $P(A)$ is the product of all parameters directly proportional with $A$, while
- $Q(A)$ the product of all parameters inversely proportional with $A$,
- $S^{\wedge}\Theta$ is the fusion space (i.e. the frame of discernment closed under union, intersection, and complement of the sets).

At the end, we normalize the result.

# 3. Unification of Filter Algorithms (UFA)

The idea is to extend the NIS [Normalized Innovation Squared] procedure and take into consideration not only the Kalman Filter (KF), Extended Kalman Filter (EKF), Unscented Kalman Filter (UKF), and Particle Filter (PF), but all filtering methods and algorithms - and use, depending on the military application and its computing complexity, the one that fits the best.

At each step, UFA should check what filter works better and use that filter.

Include linear and non-linear filters:

- Kalman Filter (KF);
- Extended Kalman Filter (EKF);
- Unscented Kalman Filter (UKF);
- Particle Filter (PF);
- Daum Filter;
- Alpha-Beta Filter;
- Alpha-Beta-Gamma Filter;
- Weiner Filter; etc.

This section is still under experimentation and research. More study and concluding examples are needed.

# 4. Unification / Combination of Image Fusion Methods (UIFM)

Firstly, we recall the definitions of T-norm and T-conorm from fuzzy logic and set, and then we present their generalizations to N-norm and N-conorm from neutrosophic logic and set.

## 4.1. T-norm and T-conorm

Defining the T-norm conjunctive consensus:

The t-norm conjunctive consensus is based on the particular t-norm function. In general, it is a function defined in fuzzy set/logic theory in order to represent the intersection between two particular fuzzy sets. If one extends T-norm to the data fusion theory, it will be a substitute for the conjunctive rule.

The T-norm has to satisfy the following conditions:

- Associativity:

$$T_{norm}\left(T_{norm}(x,y),z\right) = T_{norm}\left(x,T_{norm}(y,z)\right) \qquad (24)$$

- Commutativity:

$$T_{norm}(x,y) = T_{norm}(y,x) \qquad (25)$$

- Monotonicity:

$$if \ (x \le a) \& (y \le b) \ then \ T_{norm}(x,y) \le T_{norm}(a,b) \qquad (26)$$

- Boundary Conditions:

$$T_{norm}(x,0) = 0; \ \ T_{norm}(x,1) = x \qquad (27)$$

There are many functions which satisfy these T-norm conditions:

- Zadeh's (default) min operator:

$$m(X) = \min\{m_1(X_i), m_2(X_j)\} \tag{28}$$

- Algebraic product operator:

$$m(X) = m_1(X_i) \cdot m_2(X_j) \tag{29}$$

- Bounded product operator:

$$m(X) = \max\{0, m_1(X_i) + m_2(X_j) - 1\} \tag{30}$$

Defining the T-conorm disjunctive consensus:

The t-conorm disjunctive consensus is based on the particular t-conorm function. In general it is a function defined in fuzzy set/logic theory in order to represent the union between two particular fuzzy sets. If one extends T-conorm to the data fusion theory, it will be a substitute for the disjunctive rule.

The T-conorm has to satisfy the following conditions:

- Associativity:

$$T_{conorm}\left(T_{conorm}(x, y), z\right) = T_{conorm}\left(x, T_{conorm}(y, z)\right) \tag{31}$$

- Commutativity:

$$T_{conorm}(x, y) = T_{conorm}(y, x) \tag{32}$$

- Monotonicity:

$$if \ (x \le a) \& (y \le b) \ then \ T_{conorm}(x, y) \le T_{conorm}(a, b) \tag{33}$$

- Boundary Conditions:

$$T_{conorm}(x, 0) = x; \ \ T_{conorm}(x, 1) = 1 \tag{34}$$

There are many functions which satisfy these T-norm conditions:

- Zadeh's (default) max operator:

$$m(X) = \max\{m_1(X_i), m_2(X_j)\} \tag{35}$$

- Algebraic product operator:

$$m(X) = m_1(X_i) + m_2(X_j) - m_1(X_i) \cdot m_2(X_j) \tag{36}$$

- Bounded product operator:

$$m(X) = \min\{m_1(X_i) + m_2(X_j), 1\} \tag{37}$$

The way of association between the focal elements of the given two sources of information, $m_1(.)$ and $m_2(.)$, is defined as $X = \theta_i \cap \theta_j$ and the degree of association is as it follows:

$$\tilde{m}_{12}(X) = T_{norm}\{m_1(\theta_i) + m_2(\theta_j)\} \tag{38}$$

where $\tilde{m}_{12}(X)$ represents the basic belief assignments (*bba*) after the fusion, associated with the given proposition *X* by using particular t-norm based conjunctive rule.

## Step 2: Distribution of the mass, assigned to the conflict.

In some degree it follows the distribution of conflicting mass in the most sophisticated DSmT based Proportional Conflict Redistribution rule number 5, but the procedure here relies on fuzzy operators.

The total conflicting mass is distributed to all non-empty sets proportionally with respect to the *Maximum (Sum)* between the elements of corresponding mass matrix's columns, associated with the given element *X* of the power set. It means the bigger mass is redistributed towards the element, involved in the conflict and contributing to the conflict with the maximum specified probability mass.

The general procedure for fuzzy based conflict redistribution is as it follows:

Calculate all partial conflict masses separately;

If $\theta_i \cap \theta_j = \emptyset$, then $\theta_i$ and $\theta_j$ are involved in the conflict; redistribute the corresponding masses $m_{12}(\theta_i \cap \theta_j) > 0$, involved in the particular partial conflicts to the non-empty sets $\theta_i$ and $\theta_j$ with respect to $max\{m_1(\theta_i), m_2(\theta_j)\}$ and with respect to $max\{m_1(\theta_j), m_2(\theta_i)\}$;

Finally, for the given above two sources, the $T_{norm}$ conjunctive consensus yields:

$$\tilde{m}_{12}(\theta_i) = T_{norm}\left(m_1(\theta_i), m_2(\theta_j)\right) + T_{norm}\left(m_1(\theta_i), m_2(\theta_i \cup \theta_j)\right) + T_{norm}\left(m_1(\theta_i \cup \theta_j), m_2(\theta_i)\right) \quad (39)$$

$$\tilde{m}_{12}(\theta_j) = T_{norm}\left(m_1(\theta_j), m_2(\theta_i)\right) + T_{norm}\left(m_1(\theta_j), m_2(\theta_i \cup \theta_j)\right) + T_{norm}\left(m_1(\theta_i \cup \theta_j), m_2(\theta_j)\right) \quad (40)$$

$$\tilde{m}_{12}(\theta_i \cup \theta_j) = T_{norm}\left(m_1(\theta_i \cup \theta_j), m_2(\theta_i \cup \theta_j)\right) \quad (41)$$

## Step 3: The basic belief assignment, obtained as a result of the applied TCN rule becomes:

$$\tilde{m}_{PCR5}^{TCN}(\theta_i) = \tilde{m}_{12}(\theta_i) + m_1(\theta_i) \times \frac{T_{norm}\left(m_1(\theta_i), m_2(\theta_j)\right)}{T_{conorm}\left(m_1(\theta_i), m_2(\theta_j)\right)} +$$

$$+ m_2(\theta_i) \times \frac{T_{norm}\left(m_1(\theta_j), m_2(\theta_i)\right)}{T_{conorm}\left(m_1(\theta_j), m_2(\theta_i)\right)} \quad (42)$$

$$\tilde{m}_{PCR}^{TCN}(\theta_j) = \tilde{m}_{12}(\theta_j) + m_2(\theta_j) \times \frac{T_{norm}\left(m_1(\theta_i), m_2(\theta_j)\right)}{T_{conorm}\left(m_1(\theta_i), m_2(\theta_j)\right)} +$$

$$+ m_1(\theta_i) \times \frac{T_{norm}\left(m_1(\theta_j), m_2(\theta_i)\right)}{T_{conorm}\left(m_1(\theta_j), m_2(\theta_i)\right)} \quad (43)$$

## Step 4: Normalization of the result.

The final step of the TCN fusion rule concerns the normalization procedure:

$$\tilde{m}_{PCR5}^{TCN}(\theta) = \frac{\tilde{m}_{TCN}(\theta)}{\sum\limits_{\substack{\theta \neq \varnothing \\ \theta \in 2^\Theta}} \tilde{m}_{TCN}(\theta)} \quad (44)$$

This $\tilde{m}_{PCR5}^{TCN}(.)$ is good, but it is a deviation from PCR5 principle of distribution of the conflicting information, because it divides by $max\{m_1(A), m_2(X)\}$ and respectively by $max\{m_2(A), m_2(X)\}$. It also substitutes the conjunctive rule with the min operator.

Let's suppose $A \cap B = \emptyset$, and we have:

|       | A   | B   | $A \cup B$ | $A \cap B$ |
|-------|-----|-----|-----|-----|
| $m_1$ | 0.3 | 0   | 0.7 |     |
| $m_2$ | 0   | 0.6 | 0.4 |     |

We use the min operator for the conjunctive rule and we get:

$$\text{min}\quad 0.3 \quad 0.6 \quad 0.4 \quad 0.3$$

We need to transfer $m_{min}(A \cap B) = 0.3$ to $A$ and $B$ proportionally to their masses 0.3 and respectively 0.6.

$$\frac{x}{0.3} = \frac{y}{0.6} = \frac{\min\{0.3, 0.6\}}{0.3 + 0.6} = \frac{0.3}{0.9} = \frac{1}{3},$$

so,

$$x = 0.3\left(\frac{1}{3}\right) = 0.1, \quad y = 0.6\left(\frac{1}{3}\right) = 0.2$$

But at this step it uses the max operator and divides by $max\{...\}$.

One gets:

$$x = (0.3)\frac{\min\{0.3, 0.6\}}{\max\{0.3, 0.6\}} = 0.3 \cdot \frac{0.3}{0.6} = 0.15$$

$$y = (0.6)\frac{\min\{0.3, 0.6\}}{\max\{0.3, 0.6\}} = 0.6 \cdot \frac{0.3}{0.6} = 0.30$$

So, actually the conflict $min\{0.3, 0.6\} = 0.3 = m_{min}(A \cap B)$ is not distributed proportionally to $A$ and $B$.

A better formula (doing a better redistribution of the conflict, exactly as PCR5) is:

$$m_{PCR5v2}^{TN}(A) = \sum_{\substack{X,Y \in G^\Theta \\ X \cap Y = A}} \min\{m_1(X), m_2(Y)\} +$$

$$+ \sum_{\substack{X \in G^\Theta \\ X \cap A = \emptyset}} \left[ m_1(A) \cdot \frac{\min\{m_1(A), m_2(X)\}}{m_1(A) + m_2(X)} + m_2(A) \cdot \frac{\min\{m_2(A), m_1(X)\}}{m_2(A) + m_1(X)} \right] \quad (45)$$

And then we normalize:

$$\tilde{m}_{PCR5v2}^{TN}(A) = \frac{m_{PCR5v2}^{TN}(A)}{\sum_{X \in G^\Theta} m_{PCR5v2}^{TCN}(X)}. \quad (46)$$

We use the notation $TN$ (=T-norm) only, since T-conorm is not used.

In a more general way, we can define a class of fuzzy fusion rules based on PCR5, combining two masses $m_1(.)$ and $m_2(.)$, and corresponding to the conjunctive rule:

$$m_{PCR5v2}^{TN_g}(A) = \sum_{\substack{X,Y \in G^\Theta \\ X \cap Y = A}} TN\big(m_1(X), m_2(Y)\big) + \quad (47)$$

$$+ \sum_{\substack{X \in G^\Theta \\ X \cap A = \emptyset}} \left[ m_1(A) \cdot \frac{TN\{m_1(A), m_2(X)\}}{m_1(A) + m_2(X)} + m_2(A) \cdot \frac{TN\{m_2(A), m_1(X)\}}{m_2(A) + m_1(X)} \right],$$

where $TN(\cdot, \cdot)$ is a fuzzy T-norm. Then we normalize.

If $TN(\cdot, \cdot)$ is the product, we get just PCR5.

If $TN(\cdot, \cdot)$ is the min, we get the previous $\tilde{m}_{PCR5v2}^{TCN}(.)$.

We can replace $TN(\cdot, \cdot)$ by other fuzzy T-norms and obtain different fuzzy fusion rules.

The TN conjunctive rule is used. So,

$$m_c^{TN}(A) = \sum_{\substack{X,Y \in G^\Theta \\ X \cap Y = A}} TN\big(m_1(X), m_2(Y)\big), \quad (48)$$

and then we normalize:

$$m_D^{TN}(A) = \frac{m_c^{TN}(A)}{\sum_{X \in G^\Theta} m_c^{TN}(X)}, \quad (49)$$

which is the TN Dempster's rule.

If $TN(\cdot,\cdot)$ is the product, we get just Dempster's rule. Similarly, we can define the fuzzy disjunctive rule:

$$m_c^{TCN}(A) = \sum_{\substack{X,Y \in G^{\Theta} \\ X \cup Y = A}} TCN\big(m_1(X), m_2(Y)\big) \tag{50}$$

and then we normalize.

Thus, the TCN Dubois-Prade rule is:

$$m_{DP}^{TCN}(A) = \sum_{\substack{X,Y \in G^{\Theta} \\ X \cap Y = A}} TN\big(m_1(X), m_2(Y)\big) + \sum_{\substack{X,Y \in G^{\Theta} \\ X \cap Y = \emptyset \\ X \cup Y = A}} TCN\big(m_1(X), m_2(Y)\big) \tag{51}$$

and then we normalize.

Similarly, for DSmH, Yager's rule, Smets' rule, etc. Let's take the same example:

| | $F$ | | $H$ | | | $\emptyset$ | $\emptyset$ | $\emptyset$ |
|---|---|---|---|---|---|---|---|---|
| | $O_1$ | $O_2$ | $O_3$ | $O_1 \cup O_3$ | $O_1 \cup O_2 \cup O_3$ | $O_1 \cap O_2$ | $O_2 \cap O_3$ | $O_2 \cap (O_1 \cup O_3)$ |
| $m_x$ | 0.2 | 0.2 | 0.3 | 0 | 0.3 | | | |
| $\mu_{T_y}$ | 0 | 0.4 | 0 | 0.6 | 0 | | | |
| $m_{conj.\min}$ | 0.2 | 0.5 | 0.3 | 0.3 | 0 | 0.2 | 0.3 | 0.2 |

For TN Dempster's rule we simply normalize the masses of non-empty sets:

| | | | | | | | | |
|---|---|---|---|---|---|---|---|---|
| $TN$ Dempser's Rule | $\dfrac{2}{13}$ | $\dfrac{5}{13}$ | $\dfrac{3}{13}$ | $\dfrac{3}{13}$ | 0 | 0 | 0 | 0 |
| | | | | | | $O_1 \cup O_2$ | $O_2 \cup O_3$ | $O_2 \cup O_1 \cup O_3$ |
| $TCN$ Dubois-Prade and TCN DSmH non-normalized | 0.2 | 0.5 | 0.3 | 0.3 | 0 | 0.2 | 0.3 | 0.2 |
| | | | | | | $O_1 \cup O_2$ | $O_2 \cup O_3$ | |
| $TCN\ DP$ and TCN DSmH normalized | 0.10 | 0.25 | 0.15 | 0.15 | 0.10 | 0.10 | 0.15 | |
| $m_{PCR5v2}^{TN}$ non-normalized different formTN | 0.267 | 0.854 | 0.429 | 0.450 | 0 | 0 | 0 | 0 |

For non-norm $m_{PCR5v2}^{TCN}$ we need to transfer:

$$m_{conj.\min}(O_1 \cap O_2) = \min\{0.2, 0.4\} = 0.2$$

to $O_1$ and $O_2$ proportionally with 0.2 and 0.4 respectively (as in PCR5 ):

$$\frac{x_1}{0.2} = \frac{y_1}{0.4} = \frac{0.2}{0.2+0.4} = \frac{0.2}{0.6} = \frac{1}{3} \text{ , so } x_1 = 0.2\left(\frac{1}{3}\right) = 0.067 \text{, } y_1 = 0.4\left(\frac{1}{3}\right) = 0.133.$$

Similarly,

$$m_{conj.\min}(O_2 \cap O_3) = \min\{0.4, 0.3\} = 0.3$$

should be transferred to $O_2$ and $O_3$:

$$\frac{y_2}{0.4} = \frac{z_2}{0.3} = \frac{0.3}{0.4+0.3} = \frac{0.3}{0.7} = \frac{3}{7}; \text{ so } y_2 = 0.4\left(\frac{3}{7}\right) = 0.171 \text{, } z_2 = 0.3\left(\frac{3}{7}\right) = 0.129.$$

Same for

$$m_{conj.\min}(O_2 \cap (O_1 \cup O_3)) = \min\{0.2, 0.6\} = 0.2,$$

which should be transferred to $O_2$ and $O_1 \cup O_3$:

$$\frac{y_3}{0.2} = \frac{w_3}{0.6} = \frac{0.2}{0.2+0.6} = \frac{0.2}{0.8} = \frac{1}{4},$$

$$\text{so } y_3 = 0.2\left(\frac{1}{4}\right) = 0.05,$$

$$w_3 = 0.6\left(\frac{1}{4}\right) = 0.15$$

| | | | | | | | | |
|---|---|---|---|---|---|---|---|---|
| $m_{PCR5v2}^{TN}$ normalized different form Hebena's | 0.1335 | 0.4270 | 0.2145 | 0.2250 | 0 | 0 | 0 | 0 |
| $TN$ Yager's rule not normalized | 0.2 | 0.5 | 0.3 | 0.3 | 0.7 | | | |
| $TN$ Yager's rule normalized | 0.10 | 0.25 | 0.15 | 0.15 | 0.35 | | | |

and:

| | | | | | | Ø | |
|---|---|---|---|---|---|---|---|
| $TN$ Smet's rule not normalized | 0.2 | 0.5 | 0.3 | 0.3 | 0 | 0.7 | |
| $TN$ Smet's rule normalized | 0.10 | 0.25 | 0.15 | 0.15 | 0 | 0.35 | |

We'd define the $m_{PCR5v2}^{TCN}$ rule (with respect to PCR5) in the following condensed way:

$$m_{PCR5}^{TCN}(A) = \sum_{\substack{X,Y \in G^{\Theta} \\ X \cap Y = A}} \min\{m_1(X), m_2(Y)\} +$$

$$+ \sum_{\substack{X \in G^{\Theta} \\ X \cap A = \varnothing}} \left[ m_1(A) \cdot \frac{\min\{m_1(A), m_2(X)\}}{\max\{m_1(A) + m_2(X)\}} + m_2(A) \cdot \frac{\min\{m_2(A), m_1(X)\}}{\max\{m_2(A) + m_1(X)\}} \right]$$

and then we normalize.

The first sum is the conjunctive rule using the min operator.

We can extend these min/max operators to many other fusion rules: DSmH, Dempster's rule, disjunctive rule, Dubois-Prade's, Yager's, etc.

## 4.2. Definition of the Neutrosophic Logic/Set

Let $T, I, F$ be real standard or non-standard subsets of $]-0, 1+[$,

with
$$sup T = t\_sup, inf T = t\_inf,$$
$$sup I = i\_sup, inf I = i\_inf,$$
$$sup F = f\_sup, inf F = f\_inf,$$
and
$$n\_sup = t\_sup + i\_sup + f\_sup,$$
$$n\_inf = t\_inf + i\_inf + f\_inf.$$

Let $U$ be a universe of discourse, and $M$ a set included in $U$. An element $x$ from $U$ is noted with respect to the set $M$ as $x(T, I, F)$ and belongs to $M$ in the following way: it is $t\%$ true in the set, $i\%$ indeterminate (unknown if it is or not) in the set, and $f\%$ false, where $t$ varies in $T$, $i$ varies in $I$, $f$ varies in $F$.

Statically $T, I, F$ are subsets, but dynamically $T, I, F$ are functions/operators depending on many known or unknown parameters.

## 4.3. Neutrosophic Logic

In a similar way, we define the Neutrosophic Logic:

A logic in which each proposition $x$ is $T$% true, $I$% indeterminate, and $F$% false, and we write it $x(T, I, F)$, where $T, I, F$ are defined above.

## 4.4. N-norms and N-conorms for the Neutrosophic Logic and Set

As a generalization of T-norm and T-conorm from the Fuzzy Logic and Set, we now introduce the N-norms and N-conorms for the Neutrosophic Logic and Set.

We define a partial relation order on the neutrosophic set/logic in the following way:

$$x(T_1, I_1, F_1) \leq y(T_2, I_2, F_2)$$

if (if and only if)

$$T_1 \leq T_2, I_1 \geq I_2, F_1 \geq F_2$$

for crisp components.

And, in general, for subunitary set components:

$$x(T_1, I_1, F_1) \leq y(T_2, I_2, F_2)$$

if:

$$\inf T_1 \leq \inf T_2, \sup T_1 \leq \sup T_2,$$
$$\inf I_1 \geq \inf I_2, \sup I_1 \geq \sup I_2,$$
$$\inf F_1 \geq \inf F_2, \sup F_1 \geq \sup F_2.$$

If we have mixed - crisp and subunitary - components, or only crisp components, we can transform any crisp component, say "$a$" with $a \in [0,1]$ or $a \in ]-0, 1+[$, into a subunitary set $[a, a]$.

So, the definitions for subunitary set components should work in any case.

## 4.5. N-norms

$N_n: (\,]0,1^+[ \times\,]0,1^+[ \times\,]0,1^+[\,)^2 \rightarrow\,]0,1^+[ \times\,]0,1^+[ \times\,]0,1^+[$

$N_n\,(x(T_1,I_1,F_1),\,y(T_2,I_2,F_2)) = (N_nT(x,y),\,N_nI(x,y),\,N_nF(x,y)),$

where $N_n T(.,.), N_n I(.,.), N_n F(.,.)$ are the truth/ membership, indeterminacy, and respectively falsehood/ non-membership components. (52)

$N_n$ have to satisfy, for any $x, y, z$ in the neutrosophic logic/set $M$ of the universe of discourse $U$, the following axioms:

a) Boundary Conditions:

$N_n(x,\,\mathbf{0}) = \mathbf{0},\, N_n(x,\,\mathbf{1}) = x.$ (53)

b) Commutativity:

$N_n(x,\,y) = N_n(y,\,x).$ (54)

c) Monotonicity:

If $x \leq y$, then $N_n(x,\,z) \leq N_n(y,\,z).$ (55)

d) Associativity:

$N_n(N_n\,(x,\,y),\,z) = N_n(x,\,N_n(y,\,z)).$ (56)

There are cases when not all these axioms are satisfied, for example the associativity when dealing with the neutrosophic normalization after each neutrosophic operation. But, since we work with approximations, we can call these **N-pseudo-norms**, which still give good results in practice.

$N_n$ represent the **and** operator in neutrosophic logic, and respectively the **intersection** operator in neutrosophic set theory.

Let $J \in \{T, I, F\}$ be a component.

Most known N-norms, as in fuzzy logic and set the T-norms, are:

- The Algebraic Product N-norm:

$$\text{N}_{n-\text{algebraic}}\text{J}(x, y) = x \cdot y \tag{57}$$

- The Bounded N-Norm:

$$\text{N}_{n-\text{bounded}}\text{J}(x, y) = \max\{0, x + y - 1\} \tag{58}$$

- The Default (min) N-norm:

$$\text{N}_{n-\text{min}}\text{J}(x, y) = \min\{x, y\}. \tag{59}$$

A general example of N-norm would be this.

Let $x(T_1, I_1, F_1)$ and $y(T_2, I_2, F_2)$ be in the neutrosophic set/logic M. Then:

$$\text{N}_n(x, y) = (T_1 \wedge T_2, I_1 \vee I_2, F_1 \vee F_2) \tag{60}$$

where the " $\wedge$ " operator, acting on two (standard or non-standard) subunitary sets, is a N-norm (verifying the above N-norms axioms); while the " $\vee$ " operator, also acting on two (standard or non-standard) subunitary sets, is a N-conorm (verifying the below N-conorms axioms).

For example, $\wedge$ can be the Algebraic Product T-norm/N-norm, so $T_1 \wedge T_2 = T_1 \cdot T_2$ herein we have a product of two subunitary sets – using simplified notation); and $\vee$ can be the Algebraic Product T-conorm/N-conorm, so $T_1 \vee T_2 = T_1 + T_2 - T_1 \cdot T_2$ (herein we have a sum, then a product, and afterwards a subtraction of two subunitary sets).

Or $\wedge$ can be any T-norm/N-norm, and $\vee$ any T-conorm/N-conorm from the above and below; for example the easiest way would be to consider the min for crisp components (or inf for subset components) and respectively max for crisp components (or sup for subset components).

If we have crisp numbers, we can at the end neutrosophically normalize.

## 4.6. N-conorms

$$N_c: (\,]^-0,1^+[ \times \,]^-0,1^+[ \times \,]^-0,1^+[\,)^2 \rightarrow \,]^-0,1^+[ \times \,]^-0,1^+[ \times \,]^-0,1^+[$$

$$N_c\,(x(T_1,I_1,F_1),\,y(T_2,I_2,F_2)) = (N_cT(x,y),\,N_cI(x,y),\,N_cF(x,y)),$$

where $N_cT(.,.), N_cI(.,.), N_cF(.,.)$ are the truth/ membership, indeterminacy, and respectively falsehood/ nonmembership components. (61)

$N_c$ have to satisfy, for any $x, y, z$ in the neutrosophic logic/set $M$ of universe of discourse $U$, the following axioms:

a) Boundary Conditions:

$$N_c(x, \mathbf{1}) = \mathbf{1}, N_c(x, \mathbf{0}) = x. \tag{62}$$

b) Commutativity:

$$N_c\,(x, y) = N_c(y, x). \tag{63}$$

c) Monotonicity:

$$\text{if } x \leq y, \text{ then } N_c(x, z) \leq N_c(y, z). \tag{64}$$

d) Associativity:

$$N_c\,(N_c(x, y), z) = N_c(x, N_c(y, z)). \tag{65}$$

There are cases when not all these axioms are satisfied, for example the associativity when dealing with the neutrosophic normalization after each neutrosophic operation. But, since we work with approximations, we can call these **N-pseudo-conorms**, which still give good results in practice.

$N_c$ represent the **or** operator in neutrosophic logic, and respectively the **union** operator in neutrosophic set theory.

Let $J \in \{T, I, F\}$ be a component.

Most known N-conorms, as in fuzzy logic and set the T-conorms, are:

- The Algebraic Product N-conorm:

$$N_{c-algebraic}J(x, y) = x + y - x \cdot y \tag{66}$$

- The Bounded N-conorm:

$$N_{c-bounded}J(x, y) = \min\{1, x + y\} \tag{67}$$

- The Default (max) N-conorm:

$$N_{c-max}J(x, y) = \max\{x, y\}. \tag{68}$$

A general example of N-conorm would be this.

Let $x(T_1, I_1, F_1)$ and $y(T_2, I_2, F_2)$ be in the neutrosophic set/logic M. Then:

$$N_n(x, y) = (T_1 \vee T_2,\ I_1 \wedge I_2,\ F_1 \wedge F_2) \tag{69}$$

Where – as above - the "∧" operator, acting on two (standard or non-standard) subunitary sets, is a N-norm (verifying the above N-norms axioms); while the "∨" operator, also acting on two (standard or nonstandard) subunitary sets, is a N-conorm (verifying the above N-conorms axioms).

For example, ∧ can be the Algebraic Product T-norm/N-norm, so $T_1 \wedge T_2 = T_1 \cdot T_2$ (herein we have a product of two subunitary sets); and ∨ can be the Algebraic Product T-conorm/N-conorm, so $T_1 \vee T_2 = T_1 + T_2 - T_1 \cdot T_2$ (herein we have a sum, then a product, and afterwards a subtraction of two subunitary sets).

Or ∧ can be any T-norm/N-norm, and ∨ any T-conorm/N-conorm from the above; for example the easiest way would be to consider the min for crisp components (or inf for subset components) and respectively max for crisp components (or sup for subset components).

If we have crisp numbers, we can at the end neutrosophically normalize.

Since the min/max (or inf/sup) operators work the best for subunitary set components, let's present their definitions below. They are extensions from subunitary intervals to any subunitary sets. Analogously we can do for all neutrosophic operators.

Let $x(T_1, I_1, F_1)$ and $y(T_2, I_2, F_2)$ be in the neutrosophic set/logic M.

*Neutrosophic Conjunction/Intersection:*

$$x \wedge y = (T_\wedge, I_\wedge, F_\wedge), \tag{70}$$

where

$\inf T_\wedge = \min\{\inf T_1, \inf T_2\}$
$\sup T_\wedge = \min\{\sup T_1, \sup T_2\}$
$\inf I_\wedge = \max\{\inf I_1, \inf I_2\}$
$\sup I_\wedge = \max\{\sup I_1, \sup I_2\}$
$\inf F_\wedge = \max\{\inf F_1, \inf F_2\}$
$\sup F_\wedge = \max\{\sup F_1, \sup F_2\}$

*Neutrosophic Disjunction/Union:*

$$x \vee y = (T_\vee, I_\vee, F_\vee), \tag{71}$$

where

$\inf T_\vee = \max\{\inf T_1, \inf T_2\}$
$\sup T_\vee = \max\{\sup T_1, \sup T_2\}$
$\inf I_\vee = \min\{\inf I_1, \inf I_2\}$
$\sup I_\vee = \min\{\sup I_1, \sup I_2\}$
$\inf F_\vee = \min\{\inf F_1, \inf F_2\}$
$\sup F_\vee = \min\{\sup F_1, \sup F_2\}$

*Neutrosophic Negation/Complement:*

$$C(x) = (T_C, I_C, F_C), \tag{72}$$

where

$T_C = F_1$
$\inf I_C = 1 - \sup I_1$
$\sup I_C = 1 - \inf I_1$
$F_C = T_1$

Upon the above Neutrosophic Conjunction/ Intersection, we can define the

*Neutrosophic Containment:*

We say that the neutrosophic set $A$ is included in the neutrosophic set $B$ of the universe of discourse $U$, if for any $x(T_A, I_A, F_A) \in A$ with $x(T_B, I_B, F_B) \in B$ we have:

$$\inf T_A \leq \inf T_B \; ; \sup T_A \leq \sup T_B;$$
$$\inf I_A \geq \inf I_B \; ; \sup I_A \geq \sup I_B;$$
$$\inf F_A \geq \inf F_B \; ; \sup F_A \geq \sup F_B. \tag{73}$$

## 4.7. Remarks

a). The non-standard unit interval $]-0, 1+[$ is merely used for philosophical applications, especially when we want to make a distinction between relative truth (truth in at least one world) and absolute truth (truth in all possible worlds), and similarly for distinction between relative or absolute falsehood, and between relative or absolute indeterminacy.

But, for technical applications of neutrosophic logic and set, the domain of definition and range of the N-norm and N-conorm can be restrained to the normal standard real unit interval $[0, 1]$, which is easier to use, therefore:

$$N_n: (\,[0,1] \times [0,1] \times [0,1]\,)^2 \rightarrow [0,1] \times [0,1] \times [0,1]$$

and

$$N_c: (\,[0,1] \times [0,1] \times [0,1]\,)^2 \rightarrow [0,1] \times [0,1] \times [0,1].$$

b). Since in $NL$ and $NS$ the sum of the components (in the case when $T, I, F$ are crisp numbers, not sets) is not necessary equal to 1 (so the **normalization** is not required), we can keep the final result **un-normalized**.

But, if the normalization is needed for special applications, we can normalize at the end by dividing each component by the sum all components.

If we work with intuitionistic logic/set (when the information is incomplete, i.e. the sum of the crisp components is less than 1 , i.e. **sub-normalized**), or with paraconsistent logic/set (when the information overlaps and it is contradictory, i.e. the sum of crisp components is greater than 1 , i.e. **over-normalized**), we need to define the neutrosophic measure of a proposition/set.

If $x(T, I, F)$ is a $NL/NS$, and $T, I, F$ are crisp numbers in [0,1], then the neutrosophic vector norm of variable/set $x$ is the sum of its components:

$$N_{\text{vector-norm}}(x) = T+I+F.$$

Now, if we apply the $N_n$ and $N_c$ to two propositions/sets which maybe intuitionistic or paraconsistent or normalized (i.e. the sum of components less than 1, bigger than 1, or equal to 1), $x$ and $y$, what should be the neutrosophic measure of the results $N_n(x, y)$ and $N_c(x, y)$?

Herein again we have more possibilities:

- either the product of neutrosophic measures of $x$ and $y$:

$$N_{\text{vector-norm}}(N_n(x,y)) = N_{\text{vector-norm}}(x) \cdot N_{\text{vector-norm}}(y), \quad (74)$$

- or their average:

$$N_{\text{vector-norm}}(N_n(x,y)) = (N_{\text{vector-norm}}(x) + N_{\text{vector-norm}}(y))/2, \quad (75)$$

- or other function of the initial neutrosophic measures:

$$N_{\text{vector-norm}}(N_n(x,y)) = f(N_{\text{vector-norm}}(x), N_{\text{vector-norm}}(y)), \quad (76)$$

where $f(.,.)$ is a function to be determined according to each application.

Similarly for

$$N_{\text{vector-norm}}(N_c(x,y)).$$

Depending on the adopted neutrosophic vector norm, after applying each neutrosophic operator the result is neutrosophically normalized. We'd like to mention that "**neutrosophically normalizing**" doesn't mean that the sum of the resulting crisp components should be 1 as in fuzzy logic/set or intuitionistic fuzzy logic/set, but the sum of the components should be as above:

- either equal to the product of neutrosophic vector norms of the initial propositions/sets,
- or equal to the neutrosophic average of the initial propositions/sets vector norms, etc.

In conclusion, we neutrosophically normalize the resulting crisp components $T', I', F'$ by multiplying each neutrosophic component $T', I', F'$ with $S/(T' + I' + F')$, where

$$S = N_{vector-norm}(N_n(x,y))$$

for a N-norm or

$$S = N_{vector-norm}(N_c(x,y))$$

for a N-conorm - as defined above.

c) If $T, I, F$ are subsets of $[0, 1]$ the problem of neutrosophic normalization is more difficult.

- If $\sup(T) + \sup(I) + \sup(F) < 1$, we have an **intuitionistic proposition/set**.
- If $\inf(T) + \inf(I) + \inf(F) > 1$, we have a **paraconsistent proposition/set**.
- If there exist the crisp numbers $t \in T$, $i \in I$, and $f \in F$ such that $t + i + f = 1$, then we can say that we have a **plausible normalized proposition/set**.

But in many such cases, besides the normalized particular case showed herein, we also have crisp numbers, say $t_1 \in T$, $i_1 \in I$, and $f_1 \in F$ such that $t_1 + i_1 + f_1 < i$ (incomplete information) and $t_2 \in T$, $i_2 \in I$, and $f_2 \in F$ such that $t_2 + i_2 + f_2 > i$ (paraconsistent information).

## 4.8. Examples of Neutrosophic Operators

We define a binary **neutrosophic conjunction (intersection)** operator, which is a particular case of a N-norm (neutrosophic norm, a generalization of the fuzzy T-norm):

$$c_N^{TIF} : \left([0,1] \times [0,1] \times [0,1]\right)^2 \to [0,1] \times [0,1] \times [0,1] \quad (77)$$

$$c_N^{TIF}(x, y) = \left(T_1 T_2, I_1 I_2 + I_1 T_2 + T_1 I_2, F_1 F_2 + F_1 I_2 + F_1 T_2 + F_2 T_1 + F_2 I_1\right).$$

The neutrosophic conjunction (intersection) operator $x \wedge_N y$ component truth, indeterminacy, and falsehood values result from the multiplication

$$(T_1 + I_1 + F_1) \cdot (T_2 + I_2 + F_2) \quad (78)$$

since we consider in a prudent way $T \prec I \prec F$, where " $\prec$ " is a **neutrosophic relationship** and means "weaker", i.e. the products $T_i I_j$ will go to $I$, $T_i F_j$ will go to $F$, and $I_i F_j$ will go to $F$ for all $i, j \in \{1,2\}$, i $\neq$ j, while of course the product $T_1 T_2$ will go to $T$, $I_1 I_2$ will go to $I$, and $F_1 F_2$ will go to $F$ (or reciprocally we can say that $F$ prevails in front of $I$ which prevails in front of $T$, and this neutrosophic relationship is transitive):

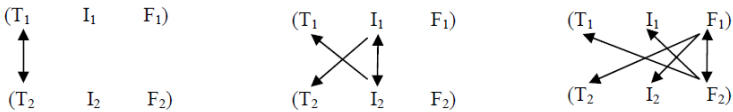

*Table 5.*

So, the truth value is $T_1 T_2$, the indeterminacy value is $I_1 I_2 + I_1 T_2 + T_1 I_2$, and the false value is $F_1 F_2 + F_1 I_2 + F_1 T_2 + F_2 T_1 + F_2 I_1$.

The norm of $x \wedge_N y$ is $(T_1 + I_1 + F_1) \cdot (T_2 + I_2 + F_2)$. Thus, if $x$ and $y$ are normalized, then $x \wedge_N y$ is also normalized. Of course, the reader can redefine the neutrosophic conjunction operator, depending on application, in a different way, for example in a more optimistic way, i.e. $I \prec T \prec F$ or $T$ prevails with respect to $I$, then we get:

$$c_N^{ITF}(x,y) = (T_1 T_2 + T_1 I_2 + T_2 I_1, I_1 I_2, F_1 F_2 + F_1 I_2 + F_1 T_2 + F_2 T_1 + F_2 I_1). \quad (79)$$

Or, the reader can consider the order $T \prec F \prec I$, etc.

Let's also define the unary neutrosophic negation operator:

$$n_N : [0,1] \times [0,1] \times [0,1] \rightarrow [0,1] \times [0,1] \times [0,1]$$

$$n_N(T,I,F) = (F,I,T) \quad (80)$$

by interchanging the truth $T$ and falsehood $F$ vector components.

Similarly, we now define a binary neutrosophic disjunction (or union) operator, where we consider the neutrosophic relationship $F \prec I \prec T$:

$$d_N^{FIT} : ([0,1] \times [0,1] \times [0,1])^2 \rightarrow [0,1] \times [0,1] \times [0,1]$$

$$d_N^{FIT}(x,y) = (T_1 T_2 + T_1 I_2 + T_2 I_1 + T_1 F_2 + T_2 F_1, I_1 I_2 + I_2 F_1 + I_1 I_2, F_1 F_2) \quad (81)$$

We consider as neutrosophic norm of the neutrosophic variable $x$, where $NL(x) = T_1 + I_1 + F_1$, the sum of its components: $T_1 + I_1 + F_1$, which in many cases is 1, but can also be positive $< 1$ or $> 1$.

Or, the reader can consider the order $F \prec T \prec I$, in a pessimistic way, i.e. focusing on indeterminacy $I$ which prevails in front of the truth $T$, or other neutrosophic order of the neutrosophic components $T, I, F$ depending on the application.

Therefore,

$$d_N^{FII}(x,y) = \left(T_1 T_2 + T_1 F_2 + T_2 F_1, I_1 F_2 + I_2 F_1 + I_1 I_2 + T_1 I_2 + T_2 I_1, F_1 F_2\right) \quad (82)$$

## 4.9. Neutrosophic Composition $k$-Law

Now, we define a more general neutrosophic composition law, named $k$-law, in order to be able to define neutrosophic $k$-conjunction/intersection and neutrosophic $k$-disjunction/union for $k$ variables, where $k \geq 2$ is an integer.

Let's consider $k \geq 2$ neutrosophic variables, $x_i(T_i + I_i + F_i$, for all $i \in \{1, 2, \ldots, k\}$. Let's denote

$$T = \left(T_1, \ldots, T_k\right)$$
$$I = \left(I_1, \ldots, I_k\right)$$
$$F = \left(F_1, \ldots, F_k\right)$$

We now define a neutrosophic composition law $o_N$ in the following way:

$$o_N : \{T, I, F\} \rightarrow [0,1]$$

If $z \in \{T, I, F\}$ then $z_{o_N} z = \prod_{i=1}^{k} z_i$ .

$$(83)$$

If $z, w \in \{T, I, F\}$ and $z \neq w$, then

$$z_{o_N} w = w_{o_N} z = \sum_{\substack{r=1 \\ \{i_1, \ldots, i_r, j_{r+1}, \ldots, j_k\} \equiv \{1,2,\ldots,k\} \\ (i_1, \ldots, i_r) \in C^r(1,2,\ldots,k) \\ (j_{r+1}, \ldots, j_k) \in C^{k-r}(1,2,\ldots,k)}}^{k-1} z_{i_1} \cdots z_{i_r} w_{j_{r+1}} \cdots w_{j_k}$$

$$(84)$$

where $C^r(1, 2, \ldots, k)$ means the set of combinations of the elements $\{1, 2, \ldots, k\}$ taken by $r$.

[Similarly for $C^{k-r}(1,2,\ldots,k)$.]

In other words, $z_{0_N}w$ is the sum of all possible products of the components of vectors $z$ and $w$, such that each product has at least a $z_i$ factor and at least a $w_j$ factor, and each product has exactly $k$ factors where each factor is a different vector component of $z$ or of $w$. Similarly if we multiply three vectors:

$$T_{0_N}I_{0_N}F = \sum_{\substack{u,v,k-u-v=1 \\ \{i_1,\ldots,i_u,j_{u+1},\ldots,j_{u+v},l_{u+v+1},\ldots,l_k\}\equiv\{1,2,\ldots,k\} \\ (i_1,\ldots,i_u)\in C^u(1,2,\ldots,k),(j_{u+1},\ldots,j_{u+v})\in \\ \in C^v(1,2,\ldots,k),(l_{u+v+1},\ldots,l_k)\in C^{k-u-v}(1,2,\ldots,k)}}^{k-2} T_{i_1\ldots i_u}I_{j_{u+1}\ldots j_{u+v}}F_{l_{u+v+1}}\ldots F_{l_k} \tag{85}$$

Let's see an example for $k = 3$.

$$x_1(T_1, I_1, F_1)$$
$$x_2(T_2, I_2, F_2)$$
$$x_3(T_3, I_3, F_3)$$

$$T_{0_N}T = T_1T_2T_3, \quad I_{0_N}I = I_1I_2I_3, \quad F_{0_N}F = F_1F_2F_3$$

$$T_{0_N}I = T_1I_2I_3 + I_1T_2I_3 + I_1I_2T_3 + T_1T_2I_3 + T_1I_2T_3 + I_1T_2T_3$$

$$T_{0_N}F = T_1F_2F_3 + F_1T_2F_3 + F_1F_2T_3 + T_1T_2F_3 + T_1F_2T_3 + F_1T_2T_3$$

$$I_{0_N}F = I_1F_2F_3 + F_1I_2F_3 + F_1F_2I_3 + I_1I_2F_3 + I_1F_2I_3 + F_1I_2I_3$$

$$T_{0_N}I_{0_N}F = T_1I_2F_3 + T_1F_2I_3 + I_1T_2F_3 + I_1F_2T_3 + F_1I_2T_3 + F_1T_2I_3$$

For the case when indeterminacy $I$ is not decomposed in subcomponents {as for example $I = P \cup U$ where $P$ =paradox (true and false simultaneously) and $U$ =uncertainty (true or false, not sure which one)}, the previous formulas can be easily written using only three components as:

$$T_{0_N}I_{0_N}F = \sum_{i,j,r\in\mathcal{P}(1,2,3)} T_iI_jF_r \tag{86}$$

where $\mathcal{P}(1,2,3)$ means the set of permutations of $(1,2,3)$ i.e.

$$\{(1,2,3),(1,3,2),(2,1,3),(2,3,1,),(3,1,2),(3,2,1)\}$$

$$z_{o_N}w = \sum_{\substack{i=1 \\ (i,j,r)\equiv(1,2,3) \\ (j,r)\in\mathcal{P}^2(1,2,3)}}^{3} z_i w_j w_{j_r} + w_i z_j z_r \tag{87}$$

This neurotrophic law is associative and commutative.

## 4.10. Neutrosophic Logic and Set $k$-Operators

Let's consider the neutrosophic logic crispy values of variables $x, y, z$ (so, for $k = 3$):

$$NL(x) = (T_1, I_1, F_1) \text{ with } 0 \le T_1, I_1, F_1 \le 1 \tag{88}$$

$$NL(y) = (T_2, I_2, F_2) \text{ with } 0 \le T_2, I_2, F_2 \le 1 \tag{89}$$

$$NL(z) = (T_3, I_3, F_3) \text{ with } 0 \le T_3, I_3, F_3 \le 1 \tag{90}$$

In neutrosophic logic it is not necessary to have the sum of components equals to 1, as in intuitionist fuzzy logic, i.e. $T_k + I_k + F_k$ is not necessary 1, for $1 \le k \le 3$.

As a particular case, we define the tri-nary conjunction neutrosophic operator:

$$c_{3_N}^{TIF} : \left([0,1]\times[0,1]\times[0,1]\right)^3 \to [0,1]\times[0,1]\times[0,1]$$

$$c_{3_N}^{TIF}(x,y,z) = \left(T_{o_N}T, I_{o_N}I + I_{o_N}T, F_{o_N}F + F_{o_N}I + F_{o_N}T\right) \tag{91}$$

If all $x, y, z$ are normalized, then $c_{3N}^{TIF}(x, y, z)$ is also normalized.

If $x, y,$ or $y$ are non-normalized, then

$$\left|c_{3_N}^{TIF}(x,y,z)\right| = |x|\cdot|y|\cdot|z| \tag{92}$$

where $|w|$ means norm of $w$.

$c_{3N}^{TIF}$ is a 3-N-norm (neutrosophic norm, i.e. generalization of the fuzzy T-norm).

Again, as a particular case, we define the unary negation neutrosophic operator:

$$n_N : [0,1] \times [0,1] \times [0,1] \rightarrow [0,1] \times [0,1] \times [0,1]$$

$$n_N(x) = n_N(T_1, I_1, F_1) = (F_1, I_1, T_1). \tag{93}$$

Let's consider the vectors:

$$T = \begin{pmatrix} T_1 \\ T_2 \\ T_3 \end{pmatrix}, \quad I = \begin{pmatrix} I_1 \\ I_2 \\ I_3 \end{pmatrix} \text{ and } F = \begin{pmatrix} F_1 \\ F_2 \\ F_3 \end{pmatrix}. \tag{94}$$

We note

$$T_x = \begin{pmatrix} F_1 \\ T_2 \\ T_3 \end{pmatrix}, \quad T_y = \begin{pmatrix} T_1 \\ F_2 \\ T_3 \end{pmatrix}, \quad T_z = \begin{pmatrix} T_1 \\ T_2 \\ F_3 \end{pmatrix}, \quad T_{xy} = \begin{pmatrix} F_1 \\ F_2 \\ T_3 \end{pmatrix}, \text{ etc.} \tag{95}$$

and similarly

$$F_x = \begin{pmatrix} T_1 \\ F_2 \\ F_3 \end{pmatrix}, \quad F_y = \begin{pmatrix} F_1 \\ T_2 \\ F_3 \end{pmatrix}, \quad F_{xz} = \begin{pmatrix} T_1 \\ F_2 \\ T_3 \end{pmatrix}, \text{ etc.} \tag{96}$$

For shorter and easier notations let's denote $z_{0_N} w = zw$ and respectively $z_{0_N} w_{0_N} = zwv$ for the vector neutrosophic law defined previously.

Then the neutrosophic tri-nary conjunction/ intersection of neutrosophic variables $x, y,$ and $z$ is:

$$c_{3N}^{TIF}(x,y,z) = (TT, II + IT, FF + FI + FT + FII) =$$
$$= (T_1 T_2 T_3, I_1 I_2 I_3 + I_1 I_2 T_3 + I_1 T_2 I_3 + T_1 I_2 I_3 + I_1 T_2 T_3 + T_1 I_2 T_3 + T_1 T_2 I_3,$$
$$F_1 F_2 F_3 + F_1 F_2 I_3 + F_1 I_2 F_3 + I_1 F_2 F_3 + F_1 I_2 I_3 + I_1 F_2 I_3 + I_1 I_2 F_3 +$$
$$+ F_1 F_2 T_3 + F_1 T_2 F_3 + T_1 F_2 F_3 + F_1 T_2 T_3 + T_1 F_2 T_3 + T_1 T_2 F_3 +$$
$$+ T_1 I_2 F_3 + T_1 F_2 I_3 + I_1 F_2 T_3 + I_1 T_2 F_3 + F_1 I_2 T_3 + F_1 T_2 I_3). \tag{97}$$

Similarly, the neutrosophic tri-nary disjunction/union of neutrosophic variables $x, y,$ and $z$ is:

$$d_{3N}^{FIT}(x,y,z) = (TT + TI + TF + TIF, II + IF, FF) =$$
$$(T_1 T_2 T_3 + T_1 T_2 I_3 + T_1 I_2 T_3 + I_1 T_2 T_3 + T_1 I_2 I_3 + I_1 I_2 T_3 + I_1 I_2 T_3 + T_1 T_2 F_3 + T_1 F_2 T_3 + F_1 T_2 T_3 +$$
$$T_1 F_2 F_3 + F_1 F_2 T_3 + F_1 F_2 T_3 + T_1 I_2 F_3 + T_1 F_2 I_3 + I_1 F_2 T_3 + I_1 T_2 F_3 + F_1 I_2 T_3 + F_1 T_2 I_3, I_1 I_2 I_3 + I_1 I_2 F_3 +$$

$$I_1F_2I_3 + F_1I_2I_3 + I_1F_2F_3 + F_1F_2I_3 + F_1F_2I_3, F_1F_2F_3) \tag{98}$$

Surely, other neutrosophic orders can be used for tri-nary conjunctions/intersections and respectively for tri-nary disjunctions/unions among the components T, I, F.

## 4.11. Neutrosophic Topologies

A). *General Definition of NT.*

Let $M$ be a non-empty set.

Let $x(T_A, I_A, F_A) \in A$ with $x(T_B, I_B, F_B) \in B$ be in the neutrosophic set/logic $M$, where $A$ and $B$ are subsets of $M$. Then (see above about N-norms/N-conorms and examples):

$$A \cup B = \{x \in M, x(T_A \lor T_B, I_A \land I_B, F_A \land F_B)\}, \tag{99}$$

$$A \cap B = \{x \in M, x(T_A \land T_B, I_A \lor I_B, F_A \lor F_B)\}, \tag{100}$$

$$C(A) = \{x \in M, x(F_A, I_A, T_A)\}. \tag{101}$$

A General Neutrosophic Topology on the non-empty set $M$ is a family η of Neutrosophic Sets in $M$ satisfying the following axioms:

$$\mathbf{0}(0,0,1) \text{ and } \mathbf{1}(1,0,0) \in \eta; \tag{102}$$

$$\text{If } A, B \in \eta, \text{ then } A \cap B \in \eta; \tag{103}$$

$$\text{If the family } \{A_k, k \in K\} \subset \eta, \text{ then } \bigcup_{k \in K} A_k \in \eta. \tag{104}$$

B). *An Alternative Version of NT.*

We cal also construct a Neutrosophic Topology on $NT = ]^-0, \ 1^+[$ considering the associated family of standard or non-standard subsets included in $NT$, and the empty set ∅, called open sets, which is closed under set union and finite intersection.

Let $A, B$ be two such subsets. The union is defined as:

$A \cup B = A + B - A \cdot B$, and the intersection as: $A \cap B = A \cdot B$. The complement of $A$, $C(A) = \{1^+\} - A$, which is a closed set. {When a non-standard number occurs at an extremity of an internal, one can write "]" instead of "(" and "[" instead of ")".} The interval *NT*, endowed with this topology, forms a **neutrosophic topological space**.

In this example, we have used the Algebraic Product N-norm/N-conorm. But other Neutrosophic Topologies can be defined by using various N-norm/N-conorm operators.

In the above defined topologies, if all $x$'s are paraconsistent or respectively intuitionistic, then one has a Neutrosophic Paraconsistent Topology, respectively Neutrosophic Intuitionistic Topology.

Much research on neutrosophic topologies has been done by Francisco Gallego Lupiañez [22].

## 4.12. Neutrosophic Logic and Set used in Image Processing

Neutrosophic logic and set have the advantage of using a third component called "indeterminacy (neutral part)", which means neither true nor false for a logical proposition, respectively neither membership nor non-membership (but unknown, unsure) of an element with respect with a set.

They are generalizations of fuzzy logic and fuzzy set, especially of intuitionistic fuzzy logic and set.

Neutrosophic logic and set have been applied in image processing thanks to their "indeterminacy" neutral component.

### 4.12.1. Removing Image Noise

H. D. Cheng and Y. Guo proposed a thresholding algorithm based on neutrosophics that will automatically select the thresholds. The thresholds are needed in order to separate the domains T and F in a neutrosophic value image.

Yanhui Guo, H. D. Cheng, and Yingtao Zhang introduced a Neutrosophic Set filter in order to denoise images. Besides pattern recognition and image vision, denoising an image is highly investigated today.

The image is converted into a neutrosophic set and then one applies a filtering method ($\gamma$-median-filtering) in order to reduce the degree of indeterminacy degree of an image; the degree of indeterminacy is found by computing the entropy of the indeterminacy subset. Afterwards the image noise is removed.

A neutrosophic image is composed of pixels, and each pixel $P$ is characterized by three components, $P_{NS}(T, I, F)$, where T=degree/percentage of truth, I=degree/percentage of indeterminacy, F=degree/percentage of falsehood.

Then a pixel $P_{NS}$ situated at the Cartesian coordinates $(i, j)$ is denoted by $P_{NS}(i, j)$.

So, we have $P_{NS}(T(i, j), I(i, j), F(i, j))$, where $T(i, j)$ is the probability that pixel $P_{NS}$ belongs to the white pixel set, $I(i, j)$ is the probability that pixel $P_{NS}$ belongs to the indeterminate pixel set, and $F(i, j)$ is the probability that pixel $P_{NS}$ belongs to the non-white pixel set. These are defined as it follows:

$$T(i, j) = \frac{\overline{g}(i, j) - \overline{g}_{\min}}{\overline{g}_{\max} - \overline{g}_{\min}} \tag{105}$$

$$\bar{g}(i,j) = \frac{1}{w \times w} \sum_{m=i-\frac{w}{2}}^{i+\frac{w}{2}} \sum_{n=j-\frac{w}{2}}^{j+\frac{w}{2}} g(m,n)$$

(106)

$$I(i,j) = \frac{\delta(i,j) - \delta_{\min}}{\delta_{\max} - \delta_{\min}}$$

(107)

$$\delta(i,j) = abs\big(g(i,j) - \bar{g}(i,j)\big)$$

(108)

$$F(i,j) = 1 - T(i,j)$$

(109)

where $\bar{g}(i,j)$ is the local mean value of the pixels of the window, while $\delta(i,j)$ is the absolute value of the difference between intensity $g(i,j)$ and its local mean value $\bar{g}(i,j)$.

**Neutrosophic image entropy** is used – for a gray image - to evaluate the distribution of the gray levels. If the intensity distribution is non-uniform, the entropy is small; but, if the intensities have equal probabilities, the entropy is high.

Neutrosophic image entropy is defined as the sum of the entropies of the three subsets T, I and F:

$$En_{NS} = En_T + En_I + En_F$$

(110)

$$En_T = -\sum_{t=\min\{T\}}^{\max\{T\}} p_T(i)\ln p_T(i)$$

(111)

$$En_I = -\sum_{t=\min\{I\}}^{\max\{I\}} p_I(i)\ln p_I(i)$$

(112)

$$En_F = -\sum_{t=\min\{F\}}^{\max\{F\}} p_F(i)\ln p_F(i)$$

(113)

where $En_T$, $En_I$ and $En_F$ are the entropies of the sets T, I and F respectively, while $p_T(i)$, $p_I(i)$, and $p_F(i)$ are the probabilities of elements in T, I and F, respectively corresponding to $i$.

In general, the median filter is known for removing the image noise in the gray level domain ($Im$). The changes in T and F influence and vary the entropy in I, which measures the indeterminacy degree of element $P_{NS}(i,j)$.

The result after median filtering, $\hat{Im}$, is defined as:

$$\hat{Im}(i,j) = \underset{(m,n)\in S_{ij}}{median}\{Im(m,n)\} \tag{114}$$

with $S_{ij}$ as the neighborhood of the pixel $(i,j)$:

Y. Guo et al. proposed the $\gamma$-median-filtering operation. A **$\gamma$-median-filtering operation** for $P_{NS}$, $\hat{P}_{NS}(\gamma)$, is defined as:

$$\hat{P}_{NS}(\gamma) = P\left(\hat{T}(\gamma), \hat{I}(\gamma), \hat{F}(\gamma)\right) \tag{115}$$

$$\hat{T}(\gamma) = \begin{cases} T & I < \gamma \\ \hat{T}_\gamma & I \geq \gamma \end{cases} \tag{116}$$

$$\hat{T}_\gamma(i,j) = \underset{(m,n)\in S_{ij}}{median}\{T(m,n)\} \tag{117}$$

$$\hat{F}(\gamma) = \begin{cases} F & I < \gamma \\ \hat{F}_\gamma & I \geq \gamma \end{cases} \tag{118}$$

$$\hat{F}_\gamma(i,j) = \underset{(m,n)\in S_{ij}}{median}\{F(m,n)\} \tag{119}$$

$$\hat{I}_\gamma(i,j) = \frac{\delta_{\hat{T}}(i,j) - \delta_{\hat{T}\min}}{\delta_{\hat{T}\max} - \delta_{\hat{T}\min}} \tag{120}$$

$$\delta_{\hat{T}}(i,j) = abs\left(\hat{T}(i,j) - \overline{\hat{T}}(i,j)\right) \tag{121}$$

$$\overline{\hat{T}}(i,j) = \frac{1}{w\times w} \sum_{m=i-\frac{w}{2}}^{i+\frac{w}{2}} \sum_{n=j-\frac{w}{2}}^{j+\frac{w}{2}} \hat{T}(m,n) \tag{122}$$

where $\delta_{\hat{T}}(i,j)$ is the absolute value of the difference between intensity $\hat{T}(i,j)$ and its local mean value $\overline{\hat{T}}(i,j)$ at $(i,j)$ after $\gamma$ -median-filtering operation.

The new neutrosophic approach to image denoising is described as below:

**Step 1:** Transform the image into $NS$ domain;

**Step 2:** Use γ-median-filtering operation on the true subset $T$ to obtain $\hat{T}_\gamma$;

**Step 3:** Compute the entropy of the indeterminate subset $\hat{I}_\gamma$, $En_{\hat{I}_\gamma}(i)$;

**Step 4:** If the following situation, go to Step 5:

$$\frac{En_{\hat{I}_\gamma}(i+1) - En_{\hat{I}_\gamma}(i)}{En_{\hat{I}_\gamma}(i)} < \delta$$

Else, $= \hat{T}_\gamma$ , go to Step 2.

**Step 5:** Transform subset $\hat{T}_\gamma$ from the neutrosophic domain into gray level domain.

The proposed method performs better for removing image noises for those noises whose types are known but also for those noises whose types are unknown.

## 4.12.2. Unification/Combination of Image Fusion Methods

Ming Zhang, Ling Zhang, H. D. Cheng used a novel approach, i.e. neutrosophic logic which is a generalization of fuzzy logic and especially of intuitionistic fuzzy logic, to image segmentation - following one of the authors (H. D. Cheng) together with his co-author Y. Guo previous published paper on neutrosophic approach to image thresholding.

The authors improved the watershed algorithms using a neutrosophic approach (i.e. they consider the objects as the T set, the background as the F set, and the edges as the I set); their method is less sensitive to noise and performs better on non-uniform images since it uses the indeterminacy (I) from neutrosophic logic and set, while this indeterminacy is not featured in fuzzy logic.

Using neutrosophic logic/set/probability/statistics is a new trend in image processing and the authors prove that the neutrosophic approach is better than other methods (such as: histogrambased, edge-based, region-based, model-based, watershed/topographic in MatLab or using Toboggan-Based).

Next step for these authors would be to use the neutrosophic approach to image registration and similarly compare the result with those obtained from other methods.

Interesting also is to use the neutrosophic approach to the control theory.

### 4.12.3. Image Segmentation

The image description, classification, and recognition depend on the image segmentation – which is used for image analysis/processing, computer vision, and pattern recognition.

Image segmentation means to find objects and boundaries such as curves, lines, etc. and partition a digital image into many regions. By image analysis, one locate objects, one measures features, one makes interpretations of scenes.

Image segmentation is done through several methods such as: histogram-based methods, region-based methods,

edge-based methods, model-based methods, and watershed methods.

The watershed uses the gradients of an image to split the image into topological areas. It best applies for uniform background and blurred edge objects, whose blurred boundaries are defined in the indeterminacy I.

After removing the noise, the image becomes more uniform.

The image is converted to a neutrosophic set in the following way.

Let $P_{\text{NS}}(i, j)$ be a pixel in the position $(i, j)$.

$$T(i,j) = \begin{cases} 0, & if\ 0 \le g_{ij} \le a; \\[2mm] \dfrac{(g_{ij} - a)^2}{(b - a)(c - a)}, & if\ a \le g_{ij} \le b; \\[4mm] 1 - \dfrac{(g_{ij} - c)^2}{(c - b)(c - a)}, & if\ b \le g_{ij} \le c; \\[2mm] 1, & if\ g_{ij} \ge c; \end{cases}$$

and $F(i, j) = q - T(i, j)$, where $g_{ij}$ is the intensity value of pixel $P(i, j)$. Then:

- Calculate the histogram of the image.
- Find the local maxima of the histogram.
- Then calculate the mean of local maxima.
- Find the peaks greater than the mean of local maxima.
- Define the low and high limits of the histogram.
- Calculate the parameters $a, b, c$ by using the maximum entropy principle: the greater the entropy is the more information the system includes.

- Find two thresholds to separate the domains T and F.
- Define homogeneity in intensity domain by using the standard deviation and the discontinuity of the intensity function. Discontinuity measures the changes in gray levels.
- Convert the image to binary image based on T, I, F.

$T(i, j)$ represents the degree of pixel $P(i, j)$ to be an object pixel;

$I(i, j)$ represents the degree of pixel $P(i, j)$ to be an edge pixel;

$F(i, j)$ represents the degree of pixel $P(i, j)$ to be a background pixel.

One determines the sets of object pixels, edge pixels, and background pixels.

- Apply the watershed for converting the binary image in the following way:

a) Get the regions $R_1, R_2, \ldots, R_n$ whose pixels are either object pixels, or edge pixels, or background pixels;

b) Dilate these regions by using the 3x3 structure element;

c) At the place where two regions merge, build a dam, until all regions merge together.

Watershed segmentation is good for uniform or nearly uniform images and the edges are connected very well. Yet, watershed method is sensitive to noise and makes over-segmentations.

# 5. Non-Linear Sequences for Target Tracking

Investigate the possibility of generalizing the result of using Fibonacci sequence (whose terms are linearly recurrent) in linear target tracking to a non-linear recurrent sequence for non-linear tracking (let us say for EKF, UKF, etc.).

Then, mathematically studying that non-linear recurrent sequence, we could get improvement of non-linear tracking.

Other non-sequences for target tracking should also be investigated and checked through concrete examples and applications.

This study has to be developed and deepened in the future.

The author has pledged in various papers, conference or seminar presentations, and scientific grant applications (between 2004-2015) for the unification of fusion theories, combinations of fusion rules, image fusion procedures, filter algorithms, and target tracking methods for more accurate applications to our real world problems - since neither fusion theory nor fusion rule fully satisfy all needed applications. For each particular application, one selects the most appropriate fusion space and fusion model, then the fusion rules, and the algorithms of implementation.

He has worked in the Unification of the Fusion Theories (UFT), which looks like a cooking recipe, better one could say like a logical chart for a computer programmer, but one does not see another method to comprise/unify all things.

The unification scenario presented herein, which is now in an incipient form, should periodically be updated incorporating new discoveries from the fusion and engineering research.